\documentclass[10pt,twocolumn,letterpaper]{article}

\usepackage[pagenumbers]{cvpr} 
\definecolor{cvprblue}{rgb}{0.21,0.49,0.74}
\usepackage[pagebackref,breaklinks,colorlinks,allcolors=cvprblue]{hyperref}
\usepackage{multirow}
\usepackage{adjustbox}
\usepackage{float}
\usepackage{pifont} 
\usepackage{stfloats}
\usepackage{subcaption}
\usepackage[rightcaption]{sidecap}
\usepackage[accsupp]{axessibility}
\def\paperID{3241} 
\def\confName{CVPR}
\def\confYear{2025}

\title{SyncVP: Joint Diffusion for Synchronous Multi-Modal Video Prediction}
\author{Enrico Pallotta, Sina Mokhtarzadeh Azar, 
Shuai Li, Olga Zatsarynna, Juergen Gall\\
University of Bonn, Lamarr Institute for Machine Learning and Artificial Intelligence\\
{\tt\small \{pallotta,mokhtarzadeh,lishuai,zatsarynna,gall\}@iai.uni-bonn.de}
}

\begin{document}
\twocolumn[{%
\renewcommand\twocolumn[1][]{#1}%
\maketitle
\vspace{-.8cm}
\begin{center}
    \centering
    \captionsetup{type=figure}
    \includegraphics[width=.95\textwidth]{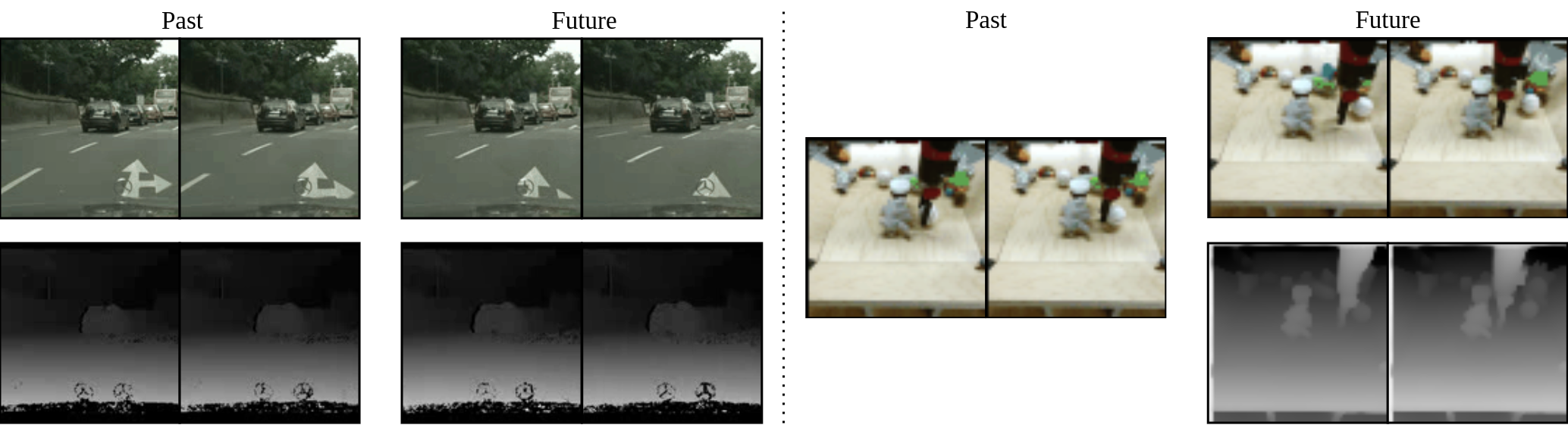}
    \captionof{figure}{SyncVP is a diffusion model for synchronized multi-modal video prediction. It generates multi-modal future frames like RGB and depth for a given observation that can consist of both modalities (left) or only one modality (right).}
    \label{fig:fig1}
\end{center}%
}]
\begin{abstract}
Predicting future video frames is essential for decision-making systems, yet RGB frames alone often lack the information needed to fully capture the underlying complexities of the real world.
To address this limitation, we propose a multi-modal framework for Synchronous Video Prediction (SyncVP) that incorporates complementary data modalities, enhancing the richness and accuracy of future predictions. SyncVP builds on pre-trained modality-specific diffusion models and introduces an efficient spatio-temporal cross-attention module to enable effective information sharing across modalities. We evaluate SyncVP on standard benchmark datasets, such as Cityscapes and BAIR, using depth as an additional modality. 
We furthermore demonstrate its generalization to other modalities on SYNTHIA with semantic information and ERA5-Land with climate data.
Notably, SyncVP achieves state-of-the-art performance, even in scenarios where only one modality is present, demonstrating its robustness and potential for a wide range of applications. 
\end{abstract}
\section{Introduction}
\label{sec:intro}
Video prediction, the task of forecasting future video frames based on past video frames, has gained significant attention in recent years~\cite{wang2017predrnn, babaeizadeh2018stochastic,tulyakov2018mocogan,MCVD} due to its broad range of applications. Autonomous driving, weather forecasting, healthcare and human-machine interaction are just a few examples of scenarios in which the ability to anticipate future events is critical. In these contexts, accurate video prediction enables systems to react and adapt in real-time, enhancing both safety and efficiency or providing valuable information for decision-making processes. 
While traditional video prediction focuses on generating future frames of RGB videos, many real-world applications involve multiple sensory inputs and require a deep understanding of the underlying real world dynamics. This leads to a natural extension of the task to the multi-modal domain, where additional data such as depth or semantic information are used alongside RGB frames. Several works have thus investigated how to exploit other modalities for conditioning the generation of images or videos, like text~\cite{Blattmann_2023_CVPR, singer2023makeavideo, zhang2023controlvideo, luo2023videofusion,Wu_2023_ICCV, guseer}, depth~\cite{xing2023make, gd-vdm, liang2024movideo, Esser_2023_ICCV, zhai2024idol}, pose~\cite{ma2024follow, zhai2024idol}, flow~\cite{ni2023conditional}, sketches~\cite{wang2024videocomposer} and audio~\cite{ruan2022mmdiffusion, xing24seeing}.
Although significant progresses have been made in guiding video generation through the use of auxiliary modalities, few studies have focused on developing models capable of leveraging multiple modalities while simultaneously generating each of them.

In this paper, we thus propose an approach for multi-modal video prediction as shown in \cref{fig:fig1}. Given a set of observed frames, the model predicts future frames of two modalities like RGB and depth in a consistent manner. While our approach is based on a latent diffusion model~\cite{PVDM}, we demonstrate that a naive concatenation of the modalities performs poorly since each modality has its own characteristics. Instead, we propose a pair of synchronized diffusion models that share relevant information through an efficient cross-attention mechanism, which works separately on the spatial and temporal dimensions. Another key aspect is the noise sharing between the modalities, which not only ensures synchronous predictions across the modalities but also has a very positive impact on the loss convergence. Finally, we propose a novel cross-modality guidance approach for training, which enables the model to predict multi-modal frames even if only one modality is observed as shown in \cref{fig:fig1}.      
We summarize the contributions of our \textit{Synchronous multi-modal Video Prediction} framework (SyncVP) in the following key points:
\begin{itemize}
    \item A generalized and scalable multi-modal framework for video prediction that can exploit pre-trained modality specific diffusion models with little finetuning. 
    \item We introduce a lightweight and efficient module for cross-modality information exchange through a split spatio-temporal cross-attention mechanism that computes only a single shared attention matrix.
    \item We propose to use a shared forward diffusion process by applying the same noise to each modality. This method leads to significant improvements in loss convergence and conditional generation performance compared to using independent noises. 
    \item We propose \textit{cross-modality guidance}, a joint modality training technique that enables simultaneous multi-modal video prediction also with partial conditioning.
    By randomly masking one or both modality inputs during training, the model learns to predict future frames even when one modality is missing. 
\end{itemize}
\section{Related Work}
\label{sec:relatedwork}
We discuss related works for video prediction and multi-modal generation, as this work bridges these two tasks and establishes a multi-modal framework for synchronous video prediction. 
\subsection{Video prediction}
Early approaches for video frames prediction primarily focused on recurrent networks like ConvLSTM and RNNs~\cite{Byeon_2018_ECCV, villegas2017decomposing, wang2017predrnn}. Due to the inherent uncertain nature of future frames in a video, several works proposed stochastic methods based on variational models (VAEs, VRNNs)~\cite{ghvaes, Castrejon_2019_ICCV, svg-lp, villegas2019high, akan2021slamp, akan2022stochastic, babaeizadeh2018stochastic, babaeizadeh2021fitvid}.
GANs have also proven effective in video prediction~\cite{clarkadversarial, tulyakov2018mocogan, luc2020transformation, lee2018savp}.
Other approaches tackled this task as a neural process~\cite{ye2023unified} mapping input spatio-temporal coordinates to target pixel values, using tailored video Transformers~\cite{ye2022vptr} or, inspired by the human vision system~\cite{stmfanet}, defining a model for the frequency domain through the use of a multi-level wavelet transform.
In the last three years, Diffusion models have been dominating the field of image generation~\cite{DDPM, latent-diff}, and they have shown outstanding performances on videos~\cite{ho2022video, Blattmann_2023_CVPR}. 
Based on the diffusion paradigm, several works proposed models for video prediction~\cite{ramvid, MCVD, STDiff, ExtDM, VDT}, each of them proposing different backbone models, past frames conditioning techniques or ways to model the spatio-temporal dependencies. 
Although these studies have progressively improved the quality of generated videos, none has yet focused on the integration and exploitation of multi-modal information.
\subsection{Multi-modal generation}
Multi-modal generation is the task of generating semantically aligned outputs across different data modalities.
In the image domain, we find several works that aim to learn the joint distribution of multiple modalities. LDM3D~\cite{stan2023ldm3d} trains a diffusion model for generating an RGB+D image from text by concatenating the RGB and depth images along the channel dimension before feeding the resulting vector to the latent encoder. MT-Diffusion~\cite{chen2024diffusion} instead defines a common diffusion space by aggregating the latent representations of each modality, which are separately encoded. DiffX~\cite{diffx} proposes a multi-path VAE to encode and decode all the modalities in a single shared latent space before training a diffusion model for layout-guidance. HyperHuman~\cite{liuhyperhuman} uses a joint diffusion UNet with expert branches of RGB, depth and surface normal for image generation. The generated depth and normal images are then used in a second refining step as conditioning for generating an RGB image with higher resolution.
Finally, Xing et al.~\cite{xing24seeing} propose an inference time optimization technique that uses a pre-trained ImageBind model to align the latent vectors generated by two separate video and audio diffusion models.

Focusing on video models, MM-Diffusion~\cite{ruan2022mmdiffusion} employs two coupled denoising networks for joint audio-video generation. The whole system is trained in a single step with a cross-attention mechanism that synchronizes the two modalities. 
Similarly, IDOL~\cite{zhai2024idol} also uses two denoising UNets coupled with cross-attention for pose-guided human image animation. They first estimate the depth map for the given image and then generate an animation of the two modalities. They furthermore introduce loss terms for motion and cross-attention maps to enhance the consistency between generated RGB and depth frames.  
Although CVD~\cite{kuang2024cvd} does not handle multiple modalities, they propose a video generative diffusion model for multiple views. A cross-view attention module is trained on top of a Stable Diffusion model to condition on different camera poses.
To the best of our knowledge, this is the first work addressing multi-modal generation in a pure video prediction manner, defining a multi-modal video-to-video framework that exploits complementary information from all the modalities to improve the generation quality of the predicted video.
\begin{figure*}[t]
  \centering
  \begin{subfigure}{0.48\textwidth}
        \centering
        \includegraphics[width=\textwidth]{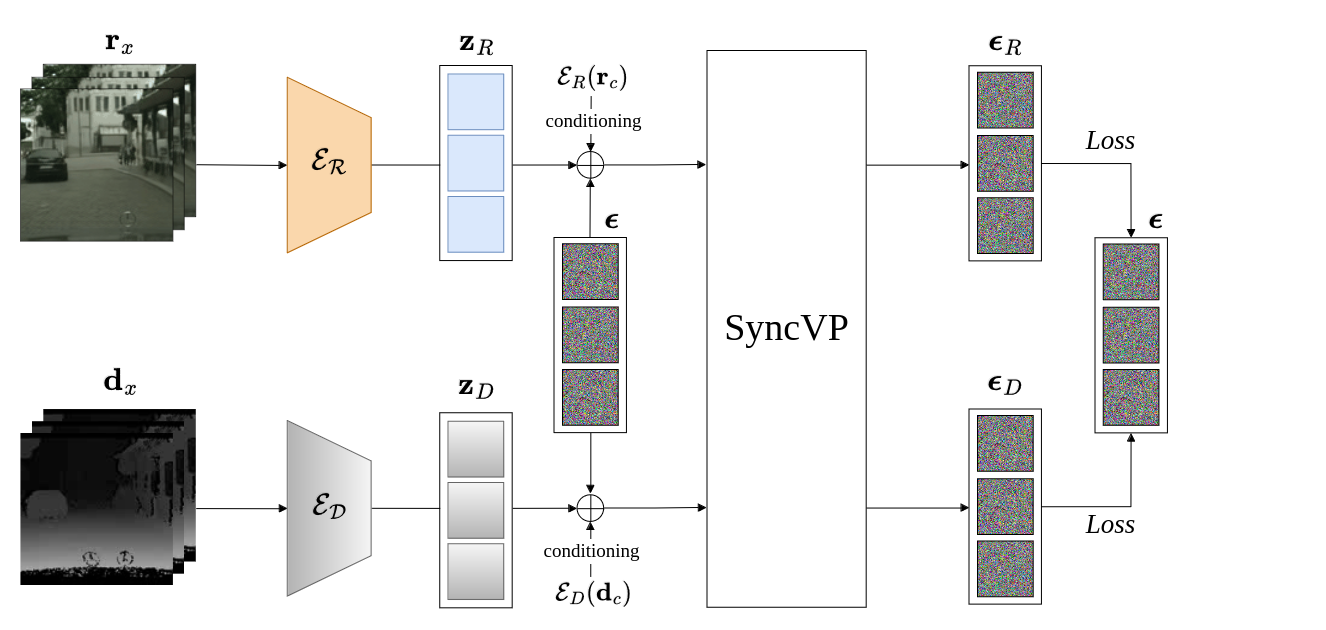}
        \caption{Joint diffusion training pipeline}
        \label{fig:sub1}
    \end{subfigure}
    \hfill
    \begin{subfigure}{0.5\textwidth}
        \centering
        \includegraphics[width=\textwidth]{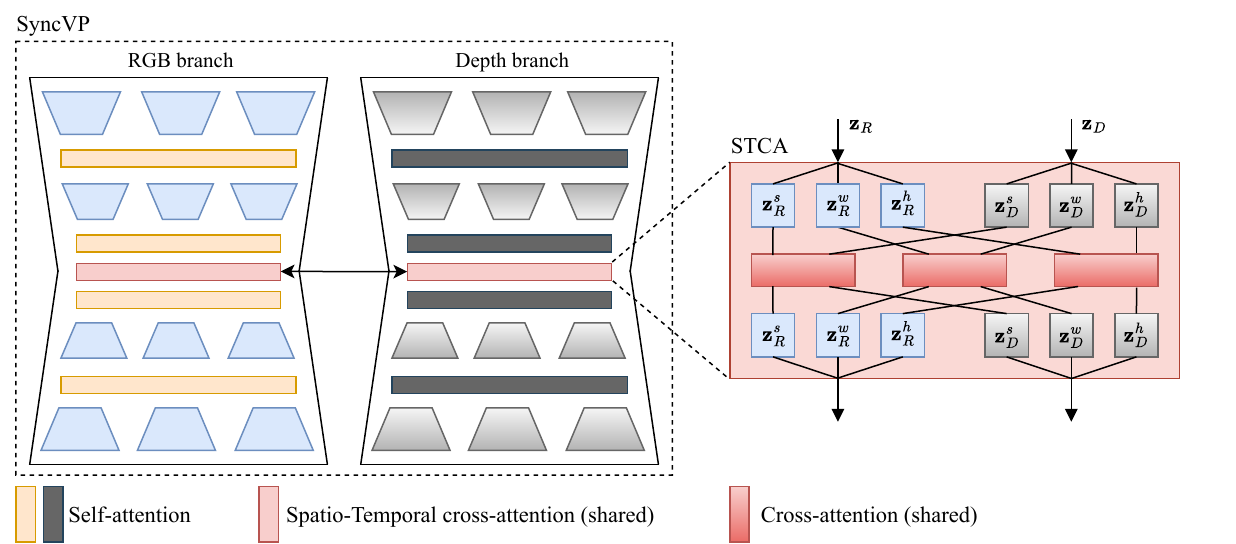}
        \caption{SyncVP architecture and cross-attention module}
        \label{fig:sub2}
    \end{subfigure}
   \caption{Our multi-modal latent diffusion framework is trained by exploiting pre-trained weights ($\theta_R,\theta_D$) of each modality and an efficient spatio-temporal cross-attention mechanism. The same noise $\boldsymbol{\epsilon}$ is used during the forward diffusion process of each modality.}
   \label{fig:model-arch}
\end{figure*}
\section{Synchronous Multi-Modal Video Prediction}
Our goal is to forecast multi-modal video frames from a short sequence of observed frames where we focus on RGB and depth as modalities. As illustrated in \cref{fig:fig1}, we propose an approach with a novel cross-modality guidance, i.e., the multi-modal model can be conditioned on both or only one modality. Before we describe the approach in detail, we first introduce a formal definition of joint multi-modal video prediction. 
Given a dataset of paired RGB and depth videos $D=\{(\mathbf{r}_i, \mathbf{d}_i)\}_{i=1}^N$, our goal is to predict $P$ future RGB $\mathbf{r}_x=(r_{t_1}, r_{t_2}, ..., r_{t_P})$ and depth frames $\mathbf{d}_x=(d_{t_1}, d_{t_2}, ..., d_{t_P})$ given $C$ past RGB $\mathbf{r}_c=(r_{t_1}, r_{t_2}, ..., r_{t_C})$ and depth frames $\mathbf{d}_c=(d_{t_1}, d_{t_2}, ..., d_{t_C})$.
We are therefore interested in learning the joint conditional distribution $p(\mathbf{r}_x,\mathbf{d}_x\mid \mathbf{r}_c,\mathbf{d}_c)$. While we describe the proposed Synchronous multi-modal Video Prediction (SyncVP) framework, which is based on a novel spatio-temporal cross-attention, in \cref{sec:network}, \cref{sec:guidance} describes the training using cross-modality guidance.
\subsection{Spatio-temporal cross-attention}\label{sec:cross-attn}
\label{sec:network}
Our proposed Synchronous multi-modal Video Prediction (SyncVP) model for RGB-D prediction consists of a latent diffusion model with two branches as illustrated in \cref{fig:sub2}. These branches are connected by a spatio-temporal cross-attention module placed between the deepest layers, right after the self-attention modules of the two denoising networks, allowing high-level semantic features alignment across the modalities. For each branch, we use a small custom version of the UNet architecture proposed by~\cite{PVDM}. The latent autoencoder takes as input a sequence of frames $\mathbf{x}\in\mathbb{R}^{T\times H\times W\times C}$ and produces a latent vector $\mathcal{E}(\mathbf{x}) = \mathbf{z} = [\mathbf{z}^s, \mathbf{z}^h, \mathbf{z}^w]\in\mathbb{R}^{C'\times L}$, where $\mathbf{z}^s\in\mathbb{R}^{C'\times\frac{H}{4}\times\frac{W}{4}}$ encodes information about the general content of the video frames, while $\mathbf{z}^h\in\mathbb{R}^{C'\times T\times \frac{H}{4}}$ and $\mathbf{z}^w\in\mathbb{R}^{C'\times T\times \frac{W}{4}}$ encode temporal information.
While one could define a simple cross-attention module that works on the intermediate features $\mathbf{z}_R = \mathcal{E}_R(\mathbf{r}_x)$ and $\mathbf{z}_D = \mathcal{E}_D(\mathbf{d}_x)$ of both UNets, we argue that a smarter and more efficient way would be to compute cross-attention only between the respective spatial and temporal latent feature vectors of the two modalities.   
Inspired by~\cite{li2021grounded, Wasim_2024_CVPR}, we propose a dual-way spatio-temporal cross-attention (STCA) that is efficiently computed by one single attention map. So given $\mathbf{z}_R = [\mathbf{z}_R^s, \mathbf{z}_R^h, \mathbf{z}_R^w]$ and $\mathbf{z}_D = [\mathbf{z}_D^s, \mathbf{z}_D^h, \mathbf{z}_D^w]$, the cross-attention is computed for each pair (3 times) of $\mathbf{z}_r\in \mathbf{z}_R$ and $\mathbf{z}_d\in \mathbf{z}_D$ as follows:
\begin{equation*}
    \begin{aligned}
        \mathbf{A} = \left( \frac{Q_R Q_D^\top}{\sqrt{d_k}} \right), \quad 
        \begin{matrix} Q_R = \mathbf{W}_{Q_R}\mathbf{z}_r, & Q_D = \mathbf{W}_{Q_D}\mathbf{z}_d, \\
        V_R = \mathbf{W}_{V_R}\mathbf{z}_r, & V_D = \mathbf{W}_{V_D}\mathbf{z}_d, 
        \end{matrix}
    \end{aligned}
\end{equation*}
\begin{equation}
    \begin{aligned}
        \mathbf{z}_r &= \mathbf{z}_r + \mathbf{W}_{O_R}(\text{Softmax}\left(\mathbf{A}\right)V_D), \\
        \mathbf{z}_d &= \mathbf{z}_d + \mathbf{W}_{O_D}(\text{Softmax}\left(\mathbf{A}^\top\right)V_R),
    \end{aligned}
    \label{eq:crossattention}
\end{equation}
where $\mathbf{W}_{Q_R}, \mathbf{W}_{Q_D}, \mathbf{W}_{V_R}, \mathbf{W}_{V_D}, \mathbf{W}_{O_R},\mathbf{ W}_{O_D}$ are the query, key and output projection matrices for RGB and depth. 
Our split Spatio-Temporal Cross-Attention (STCA) not only results in better multi-modal video predictions than using normal cross-attention (CA) as we demonstrate in the experiments, but it is also more efficient. Indeed considering the dimension of our latent vector $\mathbf{z}$ and the quadratic cost of attention, the complexity ratio between STCA and CA is:
\begin{equation}
    \frac{STCA(\mathbf{z})}{CA(\mathbf{z})}=\frac{O\left(\left( \frac{H \cdot W}{16} \right)^2 + \left( \frac{H \cdot T}{4} \right)^2 + \left( \frac{W \cdot T}{4} \right)^2\right)}
{O\left(\left(\frac{H \cdot W}{16} + \frac{H \cdot T}{4} + \frac{W \cdot T}{4}\right)^2\right)}.
\end{equation}
In case of 8 frames with resolution $64\times64$ or $128\times128$, STCA needs only $37\%$ or $50\%$ of the computation of CA, respectively. In our network, we implement STCA \eqref{eq:crossattention} as multi-head attention. 

\subsection{Cross-modality guidance}
\label{sec:guidance}
While the described architecture could be trained end-to-end to learn the joint distribution $p(\mathbf{r}_x,\mathbf{d}_x\mid \mathbf{r}_c,\mathbf{d}_c)$, such a straightforward training resulted to be ineffective. We argue that learning this distribution can be extremely complex due to the differences between the two data modalities: RGB videos contain inherently more fine grained appearance and shading details compared to depth videos. A model trained from scratch on such data would struggle to converge as each modality has its own complexity and learning curve as we show in the experiments.
To overcome this problem, we propose to first learn the two single conditional distributions $p(\mathbf{r}_x \mid \mathbf{r}_c)$ and $p(\mathbf{d}_x \mid \mathbf{d}_c)$ independently and then model the joint one in a second fine-tuning step.
We therefore train two independent diffusion models based on the PVDM~\cite{PVDM} UNet using the DDPM algorithm with the standard diffusion loss:
\begin{equation}
        \mathcal{L} = \mathbb{E}_{\mathbf{x}, \mathbf{c}, \boldsymbol{\epsilon}, t} \left[ \left\| \boldsymbol{\epsilon} - \epsilon_{\theta}(\mathcal{E}(\mathbf{x})_t, t, \mathcal{E}(\mathbf{c})) \right\|^2_2 \right],
\label{eq:single_loss}
\end{equation}
where $\theta$ are the parameters of the denoising model, $\mathbf{x}$ are the target future frames, $t$ is the diffusion timestep and $\mathbf{c}$ are the conditioning frames. 
Since we apply diffusion in a latent space, we also need to train the autoencoders for each modality: $\mathcal{D}(\mathcal{E}(\cdot))_R$ for RGB and $\mathcal{D}(\mathcal{E}(\cdot))_D$ for depth. 
After an initial step of training independently the modality specific diffusion models $\epsilon_{\theta_R}$ and $\epsilon_{\theta_D} $, we can use them as a reasonable starting point to model our joint conditional distribution $p(\mathbf{r}_x,\mathbf{d}_x \mid \mathbf{r}_c,\mathbf{d}_c)$.
So the SyncVP branches, as shown in~\cref{fig:sub2}, are initialized with these pre-trained weights and the multi-modal model is fine-tuned with respect to the following loss:
\begin{equation}
    \begin{split}
        \mathcal{L}_{M} = \mathbb{E}_{\mathbf{r}_x, \mathbf{d}_x, \mathbf{r}_c, \mathbf{d}_c, \boldsymbol{\epsilon}, t} \left[ \left\| \boldsymbol{\epsilon} - \epsilon_{\theta_R}(\mathcal{E}_R(\mathbf{r}_x)_t, t, \mathcal{E}_R(\mathbf{r}_c)) \right\|^2_2 \right. + \\
        \left.\left\| \boldsymbol{\epsilon} - \epsilon_{\theta_D}(\mathcal{E}_D(\mathbf{d}_x)_t, t, \mathcal{E}_D(\mathbf{d}_c)) \right\|^2_2 \right].
    \end{split}
\label{eq:loss}
\end{equation} 
Notice that, as illustrated in \cref{fig:sub1}, the target noise $\boldsymbol{\epsilon}$ is shared across both modalities. 
Specifically, we sample $\boldsymbol{\epsilon}\sim\mathcal{N}(\mathbf{0}, \mathbf{I})$ only once and then apply the forward diffusion process to each modality's latent vector:
\begin{equation}
    \mathbf{z}_t = \sqrt{\bar{\alpha}_t} \, \mathcal{E}(\mathbf{x}_0) + \sqrt{1 - \bar{\alpha}_t} \, \boldsymbol{\epsilon},
\label{eq:forward_diff}
\end{equation}
where $\mathbf{x}_0$ are the original input frames and $\bar{\alpha}_t = \prod_{s=1}^{t}\alpha_s$, with $\alpha_t\in(0,1)$, represents the noise schedule.
In our experiments, we show that this is particularly beneficial for video prediction and we argue that since each modality is conditioned with initial frames that belong to a shared context, learning the same reverse denoising transformation simplifies the training and enforces the model to predict frames that are consistent across modalities, leading to faster convergence, lower loss, and better conditional generation. 

Since our goal is to forecast multi-modal frames conditioned on both or only one modality as illustrated in Figure~\ref{fig:fig1}, we propose a \textit{cross-modality guidance} training procedure, which is inspired by the classifier-free guidance approach~\cite{ho2022classifier}. Instead of training the model only for the joint conditional distribution $p(\mathbf{r}_x,\mathbf{d}_x \mid \mathbf{r}_c,\mathbf{d}_c)$, we train our model for full and partial conditional generation, i.e., we simultaneously learn the following three distributions: $p(\mathbf{r}_x,\mathbf{d}_x \mid \mathbf{r}_c,\mathbf{d}_c)$, $p(\mathbf{r}_x,\mathbf{d}_x \mid \mathbf{0},\mathbf{d}_c)$, $p(\mathbf{r}_x,\mathbf{d}_x \mid \mathbf{r}_c,\mathbf{0})$. This is achieved by randomly masking one of the modalities $\mathbf{r}_c$ and $\mathbf{d}_c$.

\section{Experiments}
\subsection{Implementation details}
Analyzing a single diffusion branch, we customize the PVDM model~\cite{PVDM} as a two-level UNet, where channels are doubled, and the vector length is reduced by a factor of 4 only once. As a result, each modality-specific branch has approximately $58$ million parameters, which is about $11\%$ of the original PVDM-L model and $44\%$ of the PVDM-S model.
We train both single-modality models and SyncVP using DDPM with 1000 steps, and use DDIM~\cite{DDIM} with 100 steps during inference. 
Cross-modality guidance is applied during training by conditioning on both modalities with a $50\%$ probability, and on a single modality with a $25\%$ probability each. Our model is trained to predict 8 frames in each forward pass and operates auto-regressively at test time to generate the desired sequence length. The source code is available at \url{https://SyncVp.github.io/}.
\begin{table*}[ht]
\centering
\resizebox{0.95\linewidth}{!}{%
\begin{subtable}{0.5\textwidth}
\centering
\begin{tabular}{lcccc}
\toprule
\multirow{2}{*}{Models} & \multicolumn{4}{c}{Cityscapes, 2 $\rightarrow$ 28} \\
& \#T & FVD$\downarrow$ & SSIM$\uparrow$ & LPIPS$\downarrow$ \\
\midrule
SVG-LP~\cite{svg-lp}& 100 & 1300.26 & 0.574 & 549.0 \\
NPVP~\cite{ye2023unified} & 100 & 768.04 & 0.744 & 183.2 \\
VRNN 1L~\cite{Castrejon_2019_ICCV} & 100 & 682.08 & 0.609 & 304.0 \\
Hier-VRNN~\cite{Castrejon_2019_ICCV} & 100 & 567.51 & 0.628 & 264.0 \\
GHVAEs~\cite{ghvaes} & - & 418.00 & 0.740 & 194.0 \\
VDT~\cite{VDT} & - & 142.3 & \textbf{0.880} & - \\
MCVD~\cite{MCVD} & 10 & 141.31 & 0.690 & \underline{112.0} \\
ExtDM-K4~\cite{ExtDM} & 100 & 121.3 & \underline{0.745} & \textbf{108} \\
STDiff~\cite{STDiff} & 10 & 107.31 & 0.658 & 136.26 \\
\midrule
Ours (w/o depth) & 10 & \underline{97.31} & 0.652 & 161.1 \\
Ours & 10 & \textbf{84} & 0.649 & 159.7 \\
\bottomrule
\end{tabular}
\label{tab:citybenchmark}
\end{subtable}%
\begin{subtable}{0.5\textwidth}
\centering
\begin{tabular}{lcccc}
\toprule
\multirow{2}{*}{Models} & \multicolumn{4}{c}{BAIR, 2 $\rightarrow$ 28} \\
& \#T & FVD$\downarrow$ & SSIM$\uparrow$ & LPIPS$\downarrow$ \\
\midrule
NPVP~\cite{ye2023unified} & 100 & 923.62 & \underline{0.842} & \underline{57.43} \\
STM-FANet~\cite{stmfanet} & - & 159.6 & \textbf{0.844} & 93.6 \\
VPTR-NAR~\cite{ye2022vptr} & - & - & 0.813 & 70.0 \\
Hier-VRNN~\cite{Castrejon_2019_ICCV} & 100 & 143.4 & 0.829 & \textbf{55.0} \\
MCVD~\cite{MCVD} & 10 & 120.6 & 0.785 & 70.74 \\
SAVP~\cite{lee2018savp} & 100 & 116.4 & 0.789 & 63.4 \\
ExtDM-K4~\cite{ExtDM} & 100 & 102.8 & 0.814 & 69 \\
STDiff~\cite{STDiff} & 10 & 88.1 & 0.818 & 69.40 \\
\midrule
Ours (w/o depth) & 10 & \underline{70.49} & 0.795 & 79.43 \\
Ours & 10 & \textbf{63.60} & 0.805 & 75.17 \\
\bottomrule
\end{tabular}
\label{tab:bairbenchmark}
\end{subtable}
}
\caption{Comparison of SyncVP with other methods on Cityscapes (left) and BAIR (right).}
\label{tab:citybairbenchmark}
\end{table*}
\subsection{Datasets}
We train and evaluate our multi-modal video prediction model on two widely used datasets for video prediction, Cityscapes~\cite{Cityscapes} and BAIR~\cite{BAIR}, a subset of OpenDV-YouTube~\cite{yang2024genad}
at higher resolution,
as well as two additional datasets, SYNTHIA~\cite{SYNTHIA} and ERA5-Land~\cite{essd-13-4349-2021}, to demonstrate the generalization of the method to other modalities beyond depth.

\noindent\textbf{Cityscapes}~\cite{Cityscapes}: One of the most widely used datasets in video prediction benchmarks, Cityscapes provides RGB videos of driving scenes along with disparity maps computed from a stereo camera system mounted on a car. The dataset consists of short video sequences of 30 frames. Following previous works, we resize and center-crop the videos to a $128\times128$ resolution, using the first two frames as conditioning input to predict the remaining 28 frames ($2\rightarrow28$).

\noindent\textbf{BAIR}~\cite{BAIR}: Known for its challenging stochasticity, BAIR contains over 40.000 videos of a robotic arm making highly random movements as it interacts with objects on a tabletop.
BAIR provides only $64\times64$ RGB videos, each 30 frames in length, without depth information. To adapt this dataset to our setup, we compute pseudo ground-truth depth using DepthAnything-v2~\cite{depth_anything_v2}. Also in this case, following previous works, we adopt the $2\rightarrow28$ prediction setup.

\noindent\textbf{OpenDV-YouTube}~\cite{yang2024genad}: The dataset comprises over 1,700 hours of real-world driving videos sourced from YouTube. For our experiments, we select a small subset approximately matching the size of Cityscapes (65 minutes at 30 fps). Frames are cropped and resized to \(256 \times 256\), then these are grouped into non-overlapping clips of 32 frames. These are split into training (80\%), validation (10\%), and testing  (10\%) sets. We train this model in an $8 \rightarrow 8$ setting. Pseudo depth is computed using DepthAnything-v2~\cite{depth_anything_v2}. 

\noindent\textbf{SYNTHIA}~\cite{SYNTHIA}:
This synthetic dataset of urban scenes provides frame-by-frame aligned RGB and semantic segmentation maps. SYNTHIA consists of 182 training and 90 test video clips, each with an average length of 500 frames. For our purposes, we use 16-frame video clips at a $128\times128$ resolution, training the model in an $8 \rightarrow 8$ setting, with the test set divided into smaller, non-overlapping clips of 16 frames each. We chose to use it to test our model for RGB and semantic maps, which is a different set of modalities. 

\noindent\textbf{ERA5-Land}~\cite{essd-13-4349-2021}: The dataset is widely used in global climate research, providing daily measurements from 1950 to the present as spatial images of size $360\times540$. 
In our experiments, we focus on two variables: the two-meter temperature (t2m) and surface pressure (sp). We restrict the dataset to data from 1979 onward, yielding a total of 16,799 frames. The raw reanalysis data are sourced from the Climate Data Store (CDS)~\cite{era5process}.
Each frame is resized to $256\times384$, and the frames are then split into a training (80\%) and testing (20\%) set.
The model is trained for $4\rightarrow4$ prediction and evaluated in a $4\rightarrow8$ setup.

\subsection{Evaluation}
\paragraph{Metrics}For evaluation, we use Frechet Video Distance (FVD), Structural Similarity Index Measure (SSIM), and Learned Perceptual Image Patch Similarity (LPIPS). FVD is the primary metric, as it evaluates both temporal coherence and perceptual quality by comparing distributions between real and generated videos in feature space. SSIM assesses spatial consistency frame-by-frame, focusing on luminance and structural attributes, while LPIPS measures perceptual similarity using deep features. Unlike SSIM and LPIPS, which ignore temporal dynamics, FVD captures high-level motion and is often considered to better align with human judgment, making it particularly suitable for evaluating video prediction tasks.
Following the evaluation procedure of~\cite{STDiff, MCVD}, we also sample only 10 random future trajectories for each test sample and select the best one, as opposed to the 100 trajectories used by other methods. For a fair comparison, we report the number of trajectories (\#T) used by previous methods whenever this information is clearly stated in the paper or the evaluation code is available.
For depth and semantic segmentation, we report SSIM and $L_{2}$ error (multiplied by 100) between the generated frames and the ground-truth.

\noindent The ERA5-Land dataset evaluation is performed on 256 random samples from the test set, in this case we compute $L_1$ error for both sp and t2m.

\paragraph{Quantitative results}
We evaluate SyncVP with both RGB and depth conditioning frames on the Cityscapes and BAIR datasets in \cref{tab:citybairbenchmark}, demonstrating that our model surpasses previous state-of-the-art performance in FVD by over 21\% and 27\%, respectively.
While SSIM and LPIPS are slightly lower than optimal, they remain comparable to previous approaches, particularly those using the same number of future trajectories per test sample (\#T). We attribute the lower SSIM and LPIPS values to the significant compression applied by the autoencoder, as discussed in~\cref{sec:cross-attn}. 
Unlike prior methods that employ frame-by-frame autoencoders with a 3D latent representation, we use a spatio-temporal autoencoder that produces a simplified 2D latent vector. As shown in \cref{tab:inftime}, this compact latent structure allows our approach to outperform previous diffusion models in inference speed, achieving faster results across both single and multi-modal scenarios.

\noindent Since we leverage depth as an additional informative cue in the conditioning input, one might argue that our results are not directly comparable to prior methods that use only RGB. To address this, we also evaluate SyncVP with only RGB frames as conditioning (w/o depth) in \cref{tab:citybairbenchmark}, and demonstrate that it still achieves state-of-the-art performance. The first two rows in~\cref{fig:city_pred_without_depth} show that it consistently generates depth frames aligned with the predicted RGBs despite of missing depth data in the input.\\
For SYNTHIA (\cref{tab:synthia_results}) and ERA5-Land (\cref{tab:era5_results}), we show only a comparison between the single modality baselines and our multi-modal SyncVP, since this is the first work using these datasets for video prediction. In both cases, multi-modal training improves overall prediction performance. For ERA5-Land, surface pressure is measured in pascal (Pa) and two-meter temperature in kelvin (K).

\begin{table}[h!]
    \centering
    \resizebox{\linewidth}{!}{%
    \begin{tabular}{lcccc|cc}
    \toprule
         & MCVD~\cite{MCVD} & VDT~\cite{VDT} & ExtDM~\cite{ExtDM} & STDiff~\cite{STDiff} & RGB only & SyncVP\\
         \midrule
         Time (s) & 37.72 & 24.34 & 30.31 & 239.4 & \textbf{10.39} & \underline{22.68} \\
    \bottomrule
    \end{tabular}
    }
    \caption{Average inference time comparison for predicting 28 frames with 2 conditioning frames at $128\times128$ resolution. We use 100 sampling steps and run the models on a NVIDIA TITAN RTX GPU with batch size 1.}
    \label{tab:inftime}
\end{table}

\begin{table}[h!]
\centering
\resizebox{1.0\linewidth}{!}{%
\begin{tabular}{lccc|cc}
\toprule
\multirow{2}{*}{Models} & \multicolumn{3}{c}{RGB} & \multicolumn{2}{|c}{Sem. Segmentation} \\
& FVD$\downarrow$ & SSIM$\uparrow$ & LPIPS$\downarrow$ & SSIM$\uparrow$ & $L_{2}\downarrow$\\
\midrule
RGB & 103.81 & \textbf{0.827} & 90.29 & - & -\\
Sem. Segmentation & - & - & - & \textbf{0.649} & 9.223\\
SyncVP & \textbf{93.37} & 0.820 & \textbf{89.92} & 0.643 & \textbf{8.733}\\
\bottomrule
\end{tabular}
}
\caption{Results on SYNTHIA ($128\times128$, $8\rightarrow8$).}
\label{tab:synthia_results}
\end{table}

\begin{table}[h!]
\centering
\resizebox{0.6\linewidth}{!}{%
\begin{tabular}{lc|c}
\toprule
Models & sp (Pa) $L_1\downarrow$ & t2m (K) $L_1\downarrow$\\
\midrule
sp & 462.69 & -\\
t2m & - & 1.85\\
SyncVP & \textbf{421.65} & \textbf{1.78}\\
\bottomrule
\end{tabular}
}
\caption{Results on ERA5-Land ($256\times384$, $4\rightarrow8$).}
\label{tab:era5_results}
\end{table}

\begin{figure}[h!]
    \centering
    \includegraphics[width=1\linewidth]{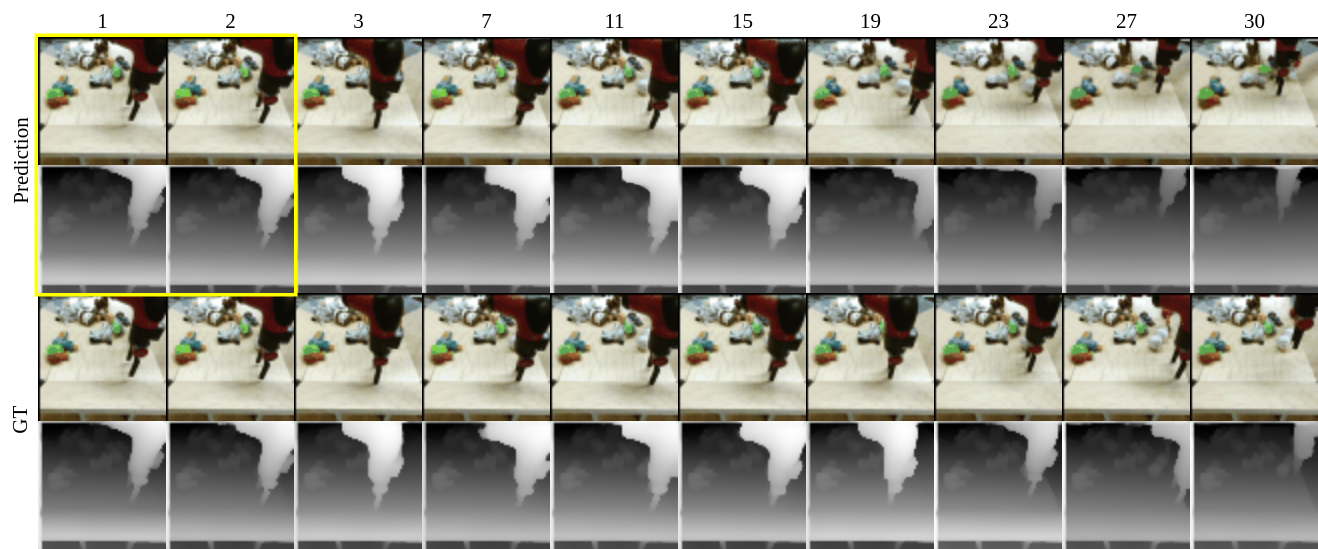}
    \caption{SyncVP predictions on BAIR using 2 conditioning frames (yellow frame). Predicting future movements of the robotic arm is challenging due to high stochasticity, but we can appreciate the alignment between predicted RGB and depth frames.}
  \label{fig:bair_pred}
\end{figure}

\vspace{-5mm}
\paragraph{Qualitative results}
We provide qualitative examples of SyncVP for the video prediction benchmarks BAIR and Cityscapes in~\cref{fig:bair_pred} and \cref{fig:city_pred_without_depth}, respectively.
\cref{fig:bair_pred} shows an example on the BAIR dataset, where the random movements of the robotic arm are unpredictable after a few frames, but the RGB and depth predictions remain well-aligned throughout the sequence. 
\cref{fig:city_pred_without_depth} is particularly important to understand the effect of our cross-modality guidance training discussed in~\cref{sec:guidance}. Without our training strategy the model is not able to predict future frames for the missing modality (row 4).

\begin{figure}[h!]
  \centering
    \includegraphics[width=1\linewidth]{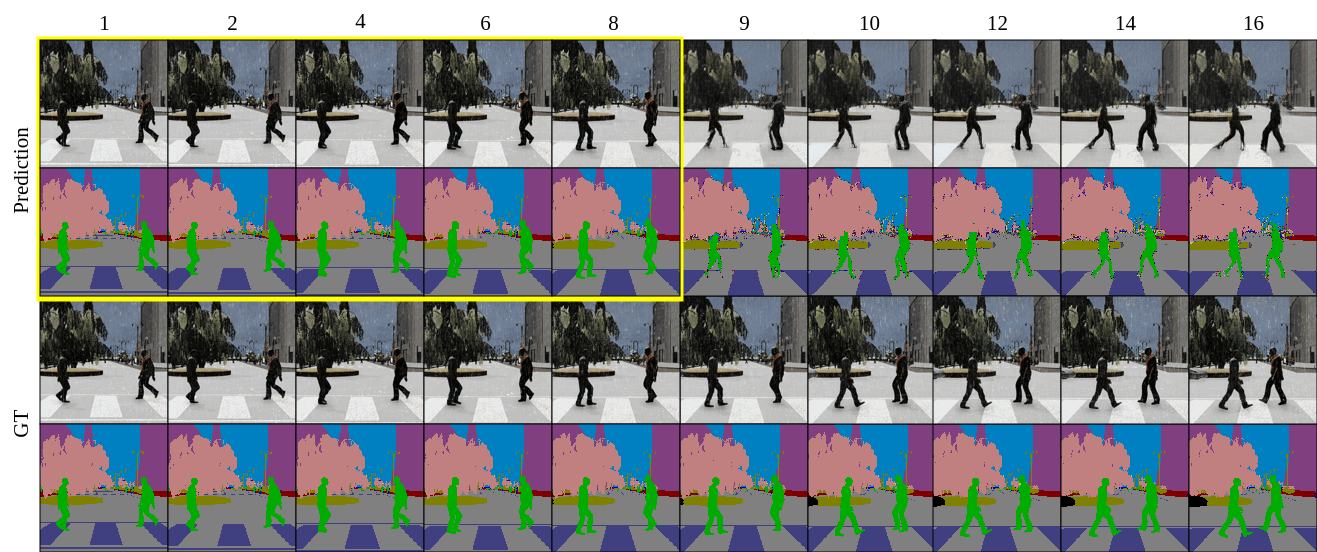}
    \caption{SyncVP predictions on SYNTHIA (RGB + Semantic segmentation) using 8 conditioning frames to predict the next 8 frames.}
  \label{fig:synthia_pred}
\end{figure}

\noindent Additional results on the SYNTHIA dataset are provided in~\cref{fig:synthia_pred}, where the colorful predicted semantic segmentation maps allow to better appreciate the alignment between the two modalities.
\cref{fig:era5_pred} shows predictions on climate data from the ERA5-Land dataset~\cite{essd-13-4349-2021}. The results show the generalization of the method to other modalities.   
In \cref{fig:youtube}, we provide some samples from the OpenDV-Youtube dataset~\cite{yang2024genad}.

\subsection{Ablations}\label{sec:ablation}
To evaluate the effectiveness of our SyncVP model for multi-modal video prediction, we conduct several experiments on the Cityscapes dataset, comparing SyncVP's performance first with single-modality models and then with two other approaches for handling multi-modal information. Inspired by LDM3D~\cite{stan2023ldm3d}, the [RGB,D] method concatenates the RGB and depth frames along the channel dimension and then trains a single autoencoder and diffusion model on them. In contrast, RGB+D with Fused Latents (FL), similar to~\cite{chen2024diffusion}, maintains separate autoencoders for each modality and fuses the latent representations by concatenating the latent vectors before applying a single diffusion model.
Additionally, we assess the efficacy of our split Spatio-Temporal Cross-Attention module (STCA) compared to vanilla cross-attention mechanism. 

\begin{figure}[h!]
  \centering
   \includegraphics[width=0.9\linewidth]{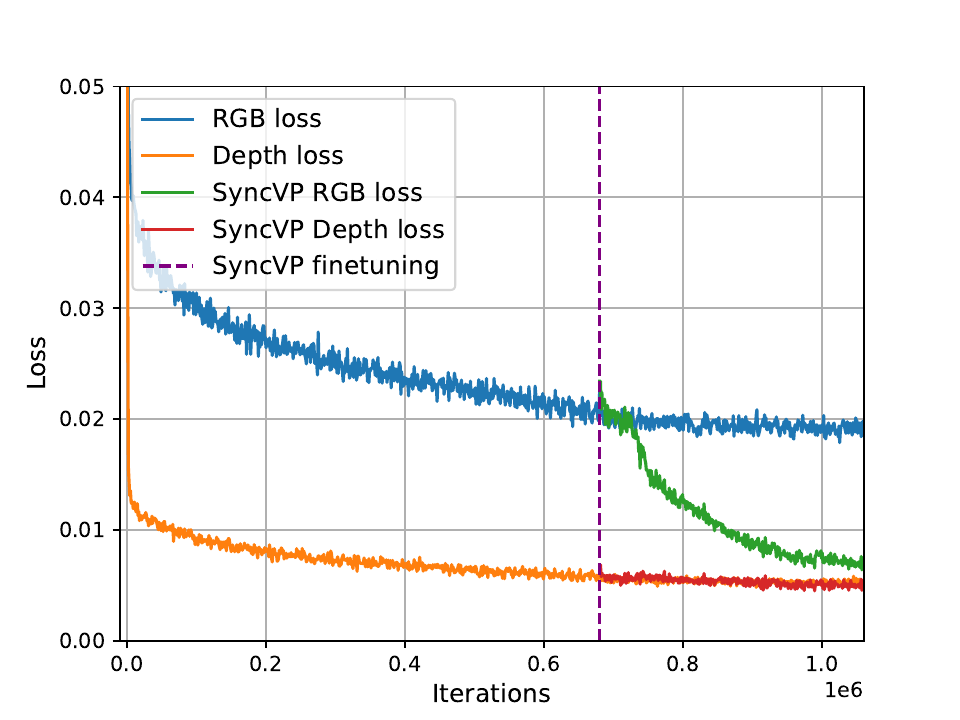}
   \caption{Comparison of RGB and depth loss for single-modality models and SyncVP on Cityscapes.}
   \label{fig:rgbd_losses}
\end{figure}

\noindent All metrics for the models mentioned above are presented in~\cref{tab:cityablations} and have been computed using models trained for the same amount of iterations: the RGB and depth models were trained for 1080k iterations, while the RGB + D (STCA) model was fine-tuned starting from the checkpoints at iteration 680k of the single-modality models  for an additional 400k iterations.
In addition to the notable improvements achieved by our SyncVP model,~\cref{fig:rgbd_losses} illustrates how training with joint modalities significantly accelerates loss reduction compared to prolonged single-modality training.
Next, we evaluate the impact of \textit{cross-modality guidance} training. Specifically,~\cref{tab:same_noise_cmg_ablation} demonstrates the benefits of this approach on video prediction quality, while  ~\cref{fig:city_pred_without_depth} shows that it enables the prediction of future frames for the missing modality, a skill that does not arise from a simple conditional training.

\noindent Finally, we assess the impact of using a shared forward diffusion process. As shown in~\cref{tab:same_noise_cmg_ablation}, training the model with independent noise for each modality results in significantly lower performance across all metrics. 
Additional experiments done while exploring different design choices are reported in \cref{tab:multiple_experiments}. The table shows that training from scratch, i.e. without pre-training the single modality models, performed poorly (row 1). Adding STCA to all layers (row
2) performed worse than adding it only to the latest layer
(row 4). Adding a motion loss (row 3) that enforces similarity between temporal latent vectors of both modalities via
an MLP projection resulted in a reduction in performance.
We also investigated the differences between using a shared
or separate STCAs (rows 5-6). Since the performance was very similar, we chose the shared STCA as it is more efficient.   
\begin{table}[h!]
\centering
\resizebox{1\linewidth}{!}{%
\begin{tabular}{lccc|cc}
\toprule
\multirow{2}{*}{Models} & \multicolumn{3}{c}{RGB} & \multicolumn{2}{|c}{Depth} \\
& FVD$\downarrow$ & SSIM$\uparrow$ & LPIPS$\downarrow$ & SSIM$\uparrow$ & $L_{2}\downarrow$\\
\midrule
RGB & 142.51 & 0.659 & 173.64 & - & -\\
Depth & - & - & - & 0.825 & 8.000\\
\relax [RGB,D] & 123.8 & 0.662 & 184.21 & 0.812 & 8.161\\
RGB + D FL & 160 & \textbf{0.667} & 180.82 & \textbf{0.835} & 7.401\\
RGB + D & 97.68 & 0.652 & 162.87 & 0.829 & 7.459\\
RGB + D STCA & \textbf{84} & 0.649 & \textbf{159.73} & 0.830 & \textbf{7.329}\\
\bottomrule
\end{tabular}
}
\caption{Ablation on Cityscapes ($128\times128$, $2\rightarrow28$). Comparing the performance of single modality baselines (rows 1 and 2) with multi-modal variants, using channel concatenation (row 3), single diffusion model with fused latents (row 4), and coupled diffusion models with vanilla (row 5) and split spatio-temporal cross attention (row 6).}
\label{tab:cityablations}
\end{table}

\begin{table}[t!]
\centering
\resizebox{1\linewidth}{!}{%
\begin{tabular}{cc|ccc|cc}
\toprule
\multirow{2}{*}{Same noise} & \multirow{2}{*}{\shortstack{Cross-modality\\guidance}} & \multicolumn{3}{c}{RGB} & \multicolumn{2}{|c}{Depth} \\
& & FVD$\downarrow$ & SSIM$\uparrow$ & LPIPS$\downarrow$ & SSIM$\uparrow$ & $L_{2}\downarrow$\\
\midrule
\ding{55} & \checkmark & 143.16 & \textbf{0.661} & 173.64 & 0.826 & 8.122\\
\checkmark & \ding{55} & 122.4 & 0.657 & 169.22 & 0.828 & 7.525\\
\checkmark & \checkmark & \textbf{84} & 0.649 & \textbf{159.73} & \textbf{0.830} & \textbf{7.329}\\
\bottomrule
\end{tabular}
}
\caption{Ablation on Cityscapes about the impact of using the same noise vs.\ independent ones (rows 1 and 3), and the impact of cross-modality guidance (rows 2 and 3).}
\label{tab:same_noise_cmg_ablation}
\end{table}

\begin{table}
\centering
\resizebox{.8\linewidth}{!}{%
\begin{tabular}{cccc}
\toprule
Variations & FVD$\downarrow$ & SSIM$\uparrow$ & LPIPS$\downarrow$ \\
\midrule
Scratch & 158.53 &\textbf{0.674} & 176.5 \\
STCA at all layers & 773.7 & 0.598 & 302.6 \\
Motion loss & 106.97 & 0.655 & 164.5 \\
\textbf{Ours} & \textbf{84} & 0.649 & \textbf{159.7} \\
\hline
Non-shared STCA$^*$  & 131.66 & 0.649 & 170.2 \\
Ours$^*$ & 129.92 & 0.650 & 171.1 \\
\bottomrule
\end{tabular}
}
\caption{Impact of training and design choices on Cityscapes.\\$^*$ denotes only 100k iterations for training.}
\label{tab:multiple_experiments}
\vspace{-0.3cm}
\end{table}

\begin{figure*}[t!]
  \centering
    \includegraphics[width=.95\linewidth]{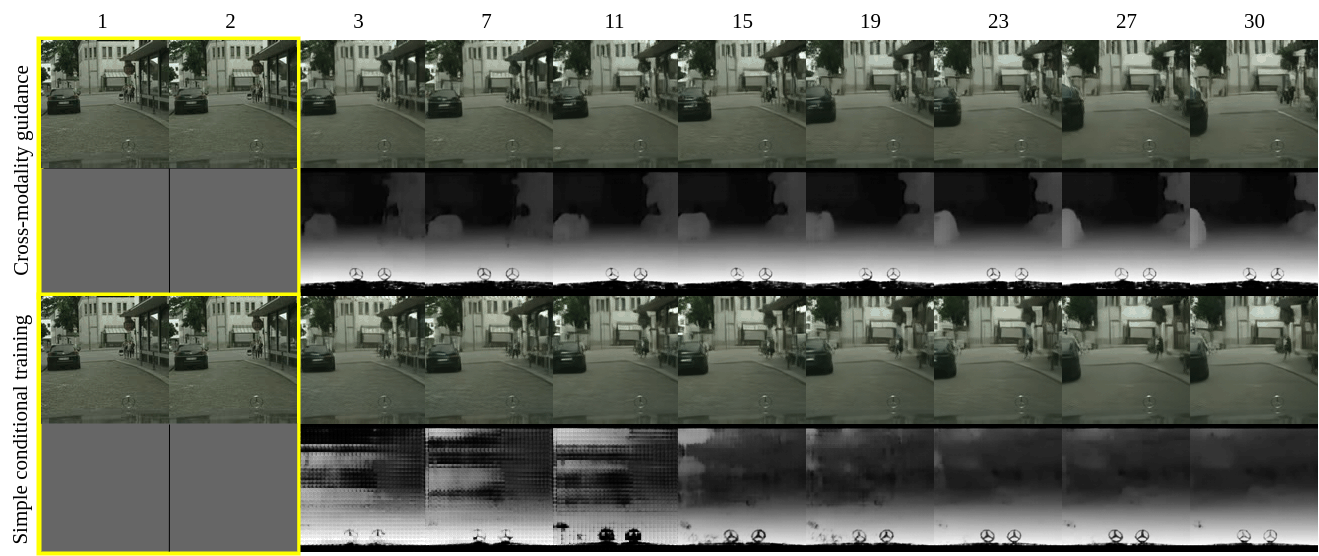}
    \caption{SyncVP predictions on Cityscapes using only RGB frames as conditioning. 
    The third and fourth row show the results if the model is trained without \textit{cross-modality guidance}, which is crucial to predict the missing modality in future frames.
    }
    \label{fig:city_pred_without_depth}
\end{figure*}

\begin{figure*}[h!]
  \centering
    \includegraphics[width=\linewidth]{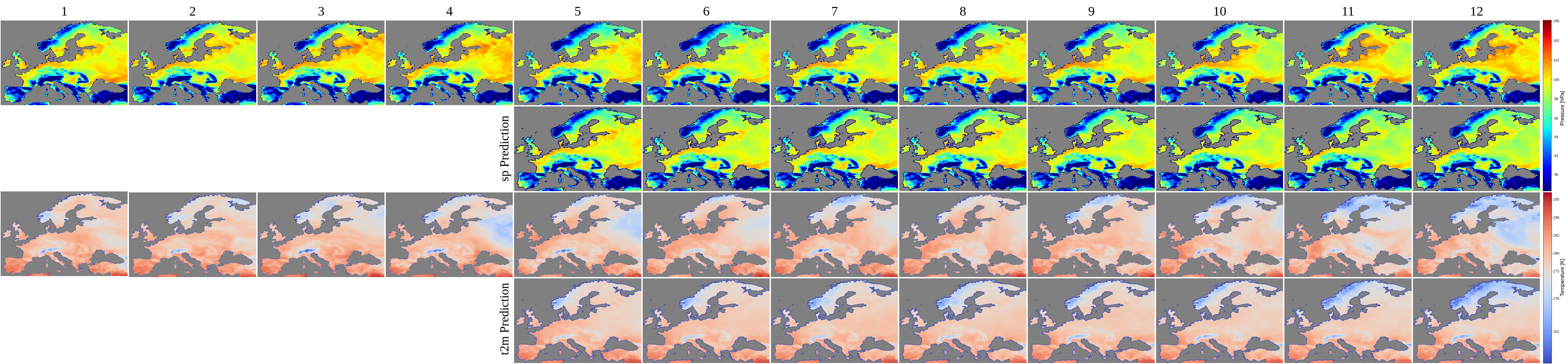}
    \caption{SyncVP predictions on ERA5-Land surface pressure (sp) and two-meter temperature (t2m) using 4 days measurements to predict the next 8 days.}
  \label{fig:era5_pred}
\end{figure*}

\begin{figure*}[h!]
  \centering
    \includegraphics[width=.95\linewidth]{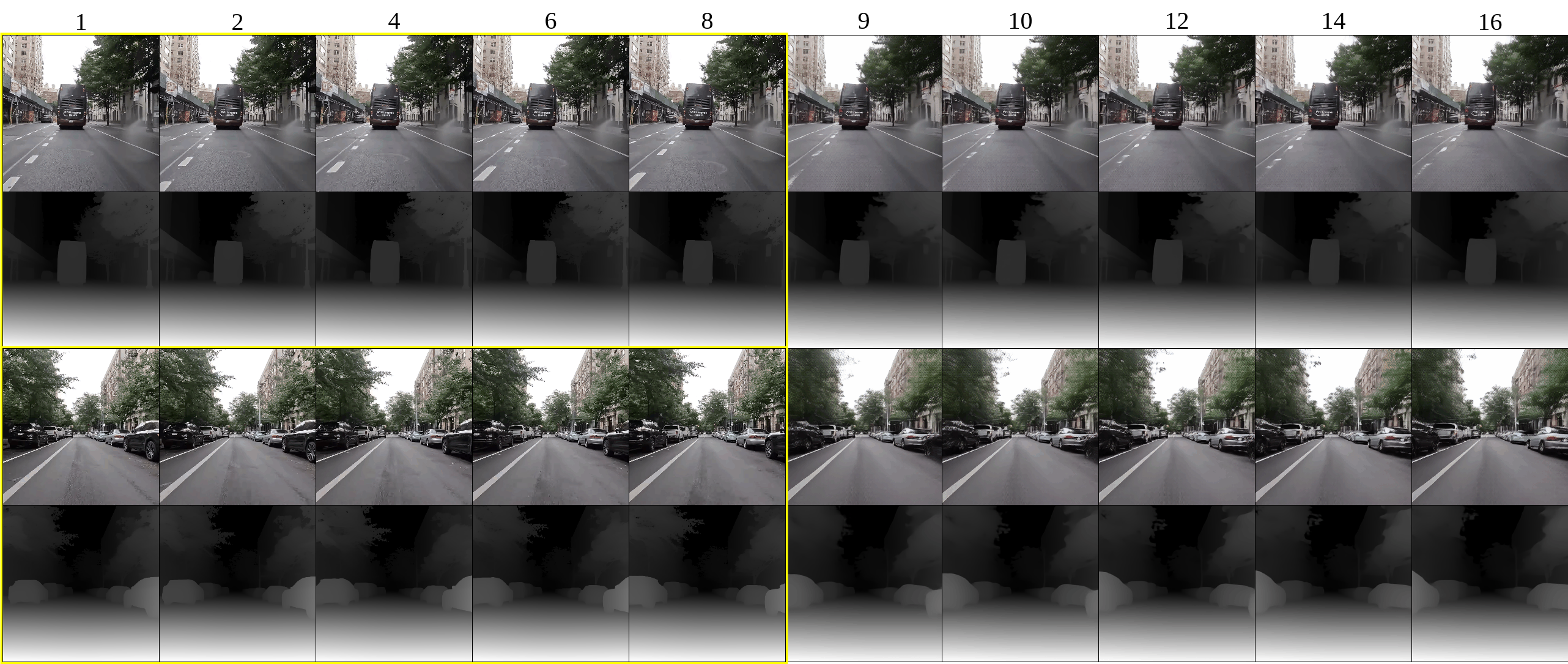}
    \caption{SyncVP predictions on OpenDV-Youtube~\cite{yang2024genad}.}
    \label{fig:youtube}
\end{figure*}
\section{Conclusions}
In this paper, we introduce SyncVP, a novel versatile framework for multi-modal video prediction. This approach is the first to leverage informative non-RGB modalities videos while efficiently predicting all target modalities in a single pass. Built upon pre-trained diffusion models, SyncVP employs a multi-branch diffusion network with spatio-temporal cross-attention to enable rich information exchange across modalities. As a result, SyncVP achieves new state-of-the-art performance on both the Cityscapes and BAIR benchmarks by over $21\%$ and $27\%$ on the main FVD metric respectively.
Notably, our model demonstrates strong predictive capability even when one of the modalities is unavailable, suggesting that jointly generating multiple modalities inherently enhances the quality of the predicted frames.
We believe that our approach can potentially pave the way to applications that require the fusion of diverse sensor inputs, a more comprehensive understanding of the predicted results and tolerance to eventual missing data.

\section*{Acknowledgements}
This work was supported by the Deutsche Forschungsgemeinschaft (DFG, German Research Foundation) GA 1927/9-1 (KI-FOR 5351), the Federal Ministry of Education and Research (BMBF) under grant no.\ 01IS24075C RAINA, and the ERC Consolidator Grant FORHUE (101044724). We sincerely acknowledge the EuroHPC Joint Undertaking for granting us access to Leonardo at CINECA, Italy, through the EuroHPC Regular Access Call (proposal No.\ EHPC-REG-2024R01-076). Additionally, the authors express their gratitude for the access provided to the Marvin cluster at the University of Bonn. We want to thank Mohamad Hakam Shams Eddin for preparing the ERA5-Land dataset.
{
    \small
    \bibliographystyle{ieeenat_fullname}
    \bibliography{main}

\begin{thebibliography}{59}
\providecommand{\natexlab}[1]{#1}
\providecommand{\url}[1]{\texttt{#1}}
\expandafter\ifx\csname urlstyle\endcsname\relax
  \providecommand{\doi}[1]{doi: #1}\else
  \providecommand{\doi}{doi: \begingroup \urlstyle{rm}\Url}\fi

\bibitem[Akan et~al.(2021)Akan, Erdem, Erdem, and G{\"u}ney]{akan2021slamp}
Adil~Kaan Akan, Erkut Erdem, Aykut Erdem, and Fatma G{\"u}ney.
\newblock Slamp: Stochastic latent appearance and motion prediction.
\newblock In \emph{Proceedings of the IEEE/CVF International Conference on Computer Vision}, pages 14728--14737, 2021.

\bibitem[Akan et~al.(2022)Akan, Safadoust, and G{\"u}ney]{akan2022stochastic}
Adil~Kaan Akan, Sadra Safadoust, and Fatma G{\"u}ney.
\newblock Stochastic video prediction with structure and motion.
\newblock \emph{arXiv preprint arXiv:2203.10528}, 2022.

\bibitem[Babaeizadeh et~al.(2018)Babaeizadeh, Finn, Erhan, Campbell, and Levine]{babaeizadeh2018stochastic}
Mohammad Babaeizadeh, Chelsea Finn, Dumitru Erhan, Roy~H Campbell, and Sergey Levine.
\newblock Stochastic variational video prediction.
\newblock In \emph{International Conference on Learning Representations}, 2018.

\bibitem[Babaeizadeh et~al.(2020)Babaeizadeh, Saffar, Nair, Levine, Finn, and Erhan]{babaeizadeh2021fitvid}
Mohammad Babaeizadeh, Mohammad~Taghi Saffar, Suraj Nair, Sergey Levine, Chelsea Finn, and Dumitru Erhan.
\newblock Fitvid: Overfitting in pixel-level video prediction.
\newblock \emph{arXiv preprint arXiv:2106.13195}, 2020.

\bibitem[Blattmann et~al.(2023)Blattmann, Rombach, Ling, Dockhorn, Kim, Fidler, and Kreis]{Blattmann_2023_CVPR}
Andreas Blattmann, Robin Rombach, Huan Ling, Tim Dockhorn, Seung~Wook Kim, Sanja Fidler, and Karsten Kreis.
\newblock Align your latents: High-resolution video synthesis with latent diffusion models.
\newblock In \emph{Proceedings of the IEEE/CVF Conference on Computer Vision and Pattern Recognition (CVPR)}, pages 22563--22575, 2023.

\bibitem[Byeon et~al.(2018)Byeon, Wang, Srivastava, and Koumoutsakos]{Byeon_2018_ECCV}
Wonmin Byeon, Qin Wang, Rupesh~Kumar Srivastava, and Petros Koumoutsakos.
\newblock Contextvp: Fully context-aware video prediction.
\newblock In \emph{Proceedings of the European Conference on Computer Vision (ECCV)}, 2018.

\bibitem[Castrejon et~al.(2019)Castrejon, Ballas, and Courville]{Castrejon_2019_ICCV}
Lluis Castrejon, Nicolas Ballas, and Aaron Courville.
\newblock Improved conditional vrnns for video prediction.
\newblock In \emph{The IEEE International Conference on Computer Vision (ICCV)}, 2019.

\bibitem[Chen et~al.(2024)Chen, Ding, Sisman, Xu, Xie, Yao, Tran, and Zeng]{chen2024diffusion}
Changyou Chen, Han Ding, Bunyamin Sisman, Yi Xu, Ouye Xie, Benjamin~Z. Yao, Son~Dinh Tran, and Belinda Zeng.
\newblock Diffusion models for multi-task generative modeling.
\newblock In \emph{The Twelfth International Conference on Learning Representations}, 2024.

\bibitem[Clark et~al.(2019)Clark, Donahue, and Simonyan]{clarkadversarial}
Aidan Clark, Jeff Donahue, and Karen Simonyan.
\newblock Adversarial video generation on complex datasets.
\newblock \emph{arXiv preprint arXiv:1907.06571}, 2019.

\bibitem[Cordts et~al.(2016)Cordts, Omran, Ramos, Rehfeld, Enzweiler, Benenson, Franke, Roth, and Schiele]{Cityscapes}
Marius Cordts, Mohamed Omran, Sebastian Ramos, Timo Rehfeld, Markus Enzweiler, Rodrigo Benenson, Uwe Franke, Stefan Roth, and Bernt Schiele.
\newblock The cityscapes dataset for semantic urban scene understanding.
\newblock In \emph{Proceedings of the IEEE Conference on Computer Vision and Pattern Recognition (CVPR)}, 2016.

\bibitem[Denton and Fergus(2018)]{svg-lp}
Emily~L. Denton and Rob Fergus.
\newblock Stochastic video generation with a learned prior.
\newblock In \emph{International Conference on Machine Learning}, 2018.

\bibitem[Ebert et~al.(2017)Ebert, Finn, Lee, and Levine]{BAIR}
Frederik Ebert, Chelsea Finn, Alex Lee, and Sergey Levine.
\newblock Self-supervised visual planning with temporal skip connections.
\newblock In \emph{Conference on Robot Learning (CoRL)}, 2017.

\bibitem[Esser et~al.(2023)Esser, Chiu, Atighehchian, Granskog, and Germanidis]{Esser_2023_ICCV}
Patrick Esser, Johnathan Chiu, Parmida Atighehchian, Jonathan Granskog, and Anastasis Germanidis.
\newblock Structure and content-guided video synthesis with diffusion models.
\newblock In \emph{Proceedings of the IEEE/CVF International Conference on Computer Vision (ICCV)}, pages 7346--7356, 2023.

\bibitem[Gu et~al.(2023)Gu, Wen, Ye, Song, and Gao]{guseer}
Xianfan Gu, Chuan Wen, Weirui Ye, Jiaming Song, and Yang Gao.
\newblock Seer: Language instructed video prediction with latent diffusion models.
\newblock In \emph{The Twelfth International Conference on Learning Representations}, 2023.

\bibitem[Ho and Salimans()]{ho2022classifier}
Jonathan Ho and Tim Salimans.
\newblock Classifier-free diffusion guidance.
\newblock In \emph{NeurIPS 2021 Workshop on Deep Generative Models and Downstream Applications}.

\bibitem[Ho et~al.(2020)Ho, Jain, and Abbeel]{DDPM}
Jonathan Ho, Ajay Jain, and Pieter Abbeel.
\newblock Denoising diffusion probabilistic models.
\newblock In \emph{Advances in Neural Information Processing Systems}, pages 6840--6851. Curran Associates, Inc., 2020.

\bibitem[Ho et~al.(2022)Ho, Salimans, Gritsenko, Chan, Norouzi, and Fleet]{ho2022video}
Jonathan Ho, Tim Salimans, Alexey Gritsenko, William Chan, Mohammad Norouzi, and David~J Fleet.
\newblock Video diffusion models.
\newblock In \emph{Advances in Neural Information Processing Systems}, pages 8633--8646. Curran Associates, Inc., 2022.

\bibitem[H{\"o}ppe et~al.(2022)H{\"o}ppe, Mehrjou, Bauer, Nielsen, and Dittadi]{ramvid}
Tobias H{\"o}ppe, Arash Mehrjou, Stefan Bauer, Didrik Nielsen, and Andrea Dittadi.
\newblock Diffusion models for video prediction and infilling.
\newblock \emph{Transactions on Machine Learning Research}, 2022.

\bibitem[Jin et~al.(2020)Jin, Hu, Tang, Niu, Shi, Han, and Li]{stmfanet}
B. Jin, Y. Hu, Q. Tang, J. Niu, Z. Shi, Y. Han, and X. Li.
\newblock Exploring spatial-temporal multi-frequency analysis for high-fidelity and temporal-consistency video prediction.
\newblock In \emph{2020 IEEE/CVF Conference on Computer Vision and Pattern Recognition (CVPR)}, pages 4553--4562, Los Alamitos, CA, USA, 2020. IEEE Computer Society.

\bibitem[Kuang et~al.(2024)Kuang, Cai, He, Xu, Li, Guibas, and Wetzstein]{kuang2024cvd}
Zhengfei Kuang, Shengqu Cai, Hao He, Yinghao Xu, Hongsheng Li, Leonidas~J Guibas, and Gordon Wetzstein.
\newblock Collaborative video diffusion: Consistent multi-video generation with camera control.
\newblock \emph{Advances in Neural Information Processing Systems}, 37:\penalty0 16240--16271, 2024.

\bibitem[Lapid et~al.(2023)Lapid, Achituve, Bracha, and Fetaya]{gd-vdm}
Ariel Lapid, Idan Achituve, Lior Bracha, and Ethan Fetaya.
\newblock Gd-vdm: Generated depth for better diffusion-based video generation.
\newblock \emph{ArXiv}, abs/2306.11173, 2023.

\bibitem[Lee et~al.(2018)Lee, Zhang, Ebert, Abbeel, Finn, and Levine]{lee2018savp}
Alex~X. Lee, Richard Zhang, Frederik Ebert, Pieter Abbeel, Chelsea Finn, and Sergey Levine.
\newblock Stochastic adversarial video prediction.
\newblock \emph{arXiv preprint arXiv:1804.01523}, 2018.

\bibitem[Li* et~al.(2022)Li*, Zhang*, Zhang*, Yang, Li, Zhong, Wang, Yuan, Zhang, Hwang, Chang, and Gao]{li2021grounded}
Liunian~Harold Li*, Pengchuan Zhang*, Haotian Zhang*, Jianwei Yang, Chunyuan Li, Yiwu Zhong, Lijuan Wang, Lu Yuan, Lei Zhang, Jenq-Neng Hwang, Kai-Wei Chang, and Jianfeng Gao.
\newblock Grounded language-image pre-training.
\newblock In \emph{CVPR}, 2022.

\bibitem[Liang et~al.(2024)Liang, Fan, Zhang, Timofte, Van~Gool, and Ranjan]{liang2024movideo}
Jingyun Liang, Yuchen Fan, Kai Zhang, Radu Timofte, Luc Van~Gool, and Rakesh Ranjan.
\newblock Movideo: Motion-aware video generation with diffusion model.
\newblock In \emph{European Conference on Computer Vision}, pages 56--74. Springer, 2024.

\bibitem[Liu et~al.(2023)Liu, Ren, Siarohin, Skorokhodov, Li, Lin, Liu, Liu, and Tulyakov]{liuhyperhuman}
Xian Liu, Jian Ren, Aliaksandr Siarohin, Ivan Skorokhodov, Yanyu Li, Dahua Lin, Xihui Liu, Ziwei Liu, and Sergey Tulyakov.
\newblock Hyperhuman: Hyper-realistic human generation with latent structural diffusion.
\newblock In \emph{The Twelfth International Conference on Learning Representations}, 2023.

\bibitem[Lu et~al.(2023)Lu, Yang, Fei, Huo, Lu, Luo, and Ding]{VDT}
Haoyu Lu, Guoxing Yang, Nanyi Fei, Yuqi Huo, Zhiwu Lu, Ping Luo, and Mingyu Ding.
\newblock Vdt: General-purpose video diffusion transformers via mask modeling.
\newblock In \emph{International Conference on Learning Representations}, 2023.

\bibitem[Luc et~al.(2020)Luc, Clark, Dieleman, Casas, Doron, Cassirer, and Simonyan]{luc2020transformation}
Pauline Luc, Aidan Clark, Sander Dieleman, Diego de~Las Casas, Yotam Doron, Albin Cassirer, and Karen Simonyan.
\newblock Transformation-based adversarial video prediction on large-scale data.
\newblock \emph{arXiv preprint arXiv:2003.04035}, 2020.

\bibitem[Luo et~al.(2023)Luo, Chen, Zhang, Huang, Wang, Shen, Zhao, Zhou, and Tan]{luo2023videofusion}
Zhengxiong Luo, Dayou Chen, Yingya Zhang, Yan Huang, Liang Wang, Yujun Shen, Deli Zhao, Jingren Zhou, and Tieniu Tan.
\newblock Videofusion: Decomposed diffusion models for high-quality video generation.
\newblock In \emph{2023 IEEE/CVF Conference on Computer Vision and Pattern Recognition (CVPR)}, pages 10209--10218. IEEE Computer Society, 2023.

\bibitem[Ma et~al.(2024)Ma, He, Cun, Wang, Chen, Li, and Chen]{ma2024follow}
Yue Ma, Yingqing He, Xiaodong Cun, Xintao Wang, Siran Chen, Xiu Li, and Qifeng Chen.
\newblock Follow your pose: Pose-guided text-to-video generation using pose-free videos.
\newblock In \emph{Proceedings of the AAAI Conference on Artificial Intelligence}, pages 4117--4125, 2024.

\bibitem[Mu\~noz Sabater(2019)]{era5process}
J. Mu\~noz Sabater.
\newblock Era5-land hourly data from 1950 to present.
\newblock \emph{Copernicus Climate Change Service (C3S) Climate Data Store (CDS)}, 2019.

\bibitem[Mu\~noz Sabater et~al.(2021)Mu\~noz Sabater, Dutra, Agust\'{\i}-Panareda, Albergel, Arduini, Balsamo, Boussetta, Choulga, and Harrigan]{essd-13-4349-2021}
J. Mu\~noz Sabater, E. Dutra, A. Agust\'{\i}-Panareda, C. Albergel, G. Arduini, G. Balsamo, S. Boussetta, M. Choulga, and S. Harrigan.
\newblock Era5-land: a state-of-the-art global reanalysis dataset for land applications.
\newblock \emph{Earth System Science Data}, 13\penalty0 (9):\penalty0 4349--4383, 2021.

\bibitem[Ni et~al.(2023)Ni, Shi, Li, Huang, and Min]{ni2023conditional}
Haomiao Ni, Changhao Shi, Kai Li, Sharon~X Huang, and Martin~Renqiang Min.
\newblock Conditional image-to-video generation with latent flow diffusion models.
\newblock In \emph{Proceedings of the IEEE/CVF Conference on Computer Vision and Pattern Recognition}, pages 18444--18455, 2023.

\bibitem[Rombach et~al.(2022)Rombach, Blattmann, Lorenz, Esser, and Ommer]{latent-diff}
Robin Rombach, Andreas Blattmann, Dominik Lorenz, Patrick Esser, and Bj\"orn Ommer.
\newblock High-resolution image synthesis with latent diffusion models.
\newblock In \emph{Proceedings of the IEEE/CVF Conference on Computer Vision and Pattern Recognition (CVPR)}, pages 10684--10695, 2022.

\bibitem[Ros et~al.(2016)Ros, Sellart, Materzynska, Vazquez, and Lopez]{SYNTHIA}
German Ros, Laura Sellart, Joanna Materzynska, David Vazquez, and Antonio~M. Lopez.
\newblock The synthia dataset: A large collection of synthetic images for semantic segmentation of urban scenes.
\newblock In \emph{2016 IEEE Conference on Computer Vision and Pattern Recognition (CVPR)}, pages 3234--3243, 2016.

\bibitem[Ruan et~al.(2023)Ruan, Ma, Yang, He, Liu, Fu, Yuan, Jin, and Guo]{ruan2022mmdiffusion}
Ludan Ruan, Yiyang Ma, Huan Yang, Huiguo He, Bei Liu, Jianlong Fu, Nicholas~Jing Yuan, Qin Jin, and Baining Guo.
\newblock Mm-diffusion: Learning multi-modal diffusion models for joint audio and video generation.
\newblock In \emph{CVPR}, 2023.

\bibitem[Singer et~al.(2023)Singer, Polyak, Hayes, Yin, An, Zhang, Hu, Yang, Ashual, Gafni, Parikh, Gupta, and Taigman]{singer2023makeavideo}
Uriel Singer, Adam Polyak, Thomas Hayes, Xi Yin, Jie An, Songyang Zhang, Qiyuan Hu, Harry Yang, Oron Ashual, Oran Gafni, Devi Parikh, Sonal Gupta, and Yaniv Taigman.
\newblock Make-a-video: Text-to-video generation without text-video data.
\newblock In \emph{The Eleventh International Conference on Learning Representations}, 2023.

\bibitem[Song et~al.(2021)Song, Meng, and Ermon]{DDIM}
Jiaming Song, Chenlin Meng, and Stefano Ermon.
\newblock Denoising diffusion implicit models.
\newblock In \emph{International Conference on Learning Representations}, 2021.

\bibitem[Stan et~al.(2023)Stan, Wofk, Fox, Redden, Saxton, Yu, Aflalo, Tseng, Nonato, Muller, et~al.]{stan2023ldm3d}
Gabriela Ben~Melech Stan, Diana Wofk, Scottie Fox, Alex Redden, Will Saxton, Jean Yu, Estelle Aflalo, Shao-Yen Tseng, Fabio Nonato, Matthias Muller, et~al.
\newblock Ldm3d: Latent diffusion model for 3d.
\newblock \emph{arXiv preprint arXiv:2305.10853}, 2023.

\bibitem[Tulyakov et~al.(2018)Tulyakov, Liu, Yang, and Kautz]{tulyakov2018mocogan}
Sergey Tulyakov, Ming-Yu Liu, Xiaodong Yang, and Jan Kautz.
\newblock Mocogan: Decomposing motion and content for video generation.
\newblock In \emph{Proceedings of the IEEE conference on computer vision and pattern recognition}, pages 1526--1535, 2018.

\bibitem[Villegas et~al.(2017)Villegas, Yang, Hong, Lin, and Lee]{villegas2017decomposing}
Ruben Villegas, Jimei Yang, Seunghoon Hong, Xunyu Lin, and Honglak Lee.
\newblock Decomposing motion and content for natural video sequence prediction.
\newblock In \emph{International Conference on Learning Representations}, 2017.

\bibitem[Villegas et~al.(2019)Villegas, Pathak, Kannan, Erhan, Le, and Lee]{villegas2019high}
Ruben Villegas, Arkanath Pathak, Harini Kannan, Dumitru Erhan, Quoc~V Le, and Honglak Lee.
\newblock High fidelity video prediction with large stochastic recurrent neural networks.
\newblock \emph{Advances in Neural Information Processing Systems}, 32, 2019.

\bibitem[Voleti et~al.(2022)Voleti, Jolicoeur-Martineau, and Pal]{MCVD}
Vikram Voleti, Alexia Jolicoeur-Martineau, and Christopher Pal.
\newblock Mcvd: Masked conditional video diffusion for prediction, generation, and interpolation.
\newblock In \emph{(NeurIPS) Advances in Neural Information Processing Systems}, 2022.

\bibitem[Wang et~al.(2024{\natexlab{a}})Wang, Yuan, Zhang, Chen, Wang, Zhang, Shen, Zhao, and Zhou]{wang2024videocomposer}
Xiang Wang, Hangjie Yuan, Shiwei Zhang, Dayou Chen, Jiuniu Wang, Yingya Zhang, Yujun Shen, Deli Zhao, and Jingren Zhou.
\newblock Videocomposer: Compositional video synthesis with motion controllability.
\newblock \emph{Advances in Neural Information Processing Systems}, 36, 2024{\natexlab{a}}.

\bibitem[Wang et~al.(2017)Wang, Long, Wang, Gao, and Yu]{wang2017predrnn}
Yunbo Wang, Mingsheng Long, Jianmin Wang, Zhifeng Gao, and Philip~S Yu.
\newblock Predrnn: Recurrent neural networks for predictive learning using spatiotemporal lstms.
\newblock \emph{Advances in neural information processing systems}, 30, 2017.

\bibitem[Wang et~al.(2024{\natexlab{b}})Wang, Lin, Qian, Huang, Tian, Chai, Deng, Du, Chen, Guo, and Huang]{diffx}
Zeyu Wang, Jingyu Lin, Yifei Qian, Yi Huang, Shicen Tian, Bosong Chai, Juncan Deng, Lan Du, Cunjian Chen, Yufei Guo, and Kejie Huang.
\newblock Diffx: Guide your layout to cross-modal generative modeling.
\newblock \emph{ArXiv}, abs/2407.15488, 2024{\natexlab{b}}.

\bibitem[Wasim et~al.(2024)Wasim, Naseer, Khan, Yang, and Khan]{Wasim_2024_CVPR}
Syed~Talal Wasim, Muzammal Naseer, Salman Khan, Ming-Hsuan Yang, and Fahad~Shahbaz Khan.
\newblock Videogrounding-dino: Towards open-vocabulary spatio-temporal video grounding.
\newblock In \emph{Proceedings of the IEEE/CVF Conference on Computer Vision and Pattern Recognition (CVPR)}, pages 18909--18918, 2024.

\bibitem[Wu et~al.(2021)Wu, Nair, Martin-Martin, Fei-Fei, and Finn]{ghvaes}
B. Wu, S. Nair, R. Martin-Martin, L. Fei-Fei, and C. Finn.
\newblock Greedy hierarchical variational autoencoders for large-scale video prediction.
\newblock In \emph{2021 IEEE/CVF Conference on Computer Vision and Pattern Recognition (CVPR)}, pages 2318--2328, Los Alamitos, CA, USA, 2021. IEEE Computer Society.

\bibitem[Wu et~al.(2023)Wu, Ge, Wang, Lei, Gu, Shi, Hsu, Shan, Qie, and Shou]{Wu_2023_ICCV}
Jay~Zhangjie Wu, Yixiao Ge, Xintao Wang, Stan~Weixian Lei, Yuchao Gu, Yufei Shi, Wynne Hsu, Ying Shan, Xiaohu Qie, and Mike~Zheng Shou.
\newblock Tune-a-video: One-shot tuning of image diffusion models for text-to-video generation.
\newblock In \emph{Proceedings of the IEEE/CVF International Conference on Computer Vision (ICCV)}, pages 7623--7633, 2023.

\bibitem[Xing et~al.(2023)Xing, Xia, Liu, Zhang, Zhang, He, Liu, Chen, Cun, Wang, Shan, and Wong]{xing2023make}
Jinbo Xing, Menghan Xia, Yuxin Liu, Yuechen Zhang, Yong Zhang, Yin-Yin He, Hanyuan Liu, Haoxin Chen, Xiaodong Cun, Xintao Wang, Ying Shan, and Tien-Tsin Wong.
\newblock Make-your-video: Customized video generation using textual and structural guidance.
\newblock \emph{IEEE transactions on visualization and computer graphics}, PP, 2023.

\bibitem[Xing et~al.(2024)Xing, He, Tian, Wang, and Chen]{xing24seeing}
Yazhou Xing, Yingqing He, Zeyue Tian, Xintao Wang, and Qifeng Chen.
\newblock Seeing and hearing: Open-domain visual-audio generation with diffusion latent aligners.
\newblock In \emph{CVPR}, 2024.

\bibitem[Yang et~al.(2024{\natexlab{a}})Yang, Gao, Qiu, Chen, Li, Dai, Chitta, Wu, Zeng, Luo, Zhang, Geiger, Qiao, and Li]{yang2024genad}
Jiazhi Yang, Shenyuan Gao, Yihang Qiu, Li Chen, Tianyu Li, Bo Dai, Kashyap Chitta, Penghao Wu, Jia Zeng, Ping Luo, Jun Zhang, Andreas Geiger, Yu Qiao, and Hongyang Li.
\newblock Generalized predictive model for autonomous driving.
\newblock In \emph{Proceedings of the IEEE/CVF Conference on Computer Vision and Pattern Recognition}, 2024{\natexlab{a}}.

\bibitem[Yang et~al.(2024{\natexlab{b}})Yang, Kang, Huang, Zhao, Xu, Feng, and Zhao]{depth_anything_v2}
Lihe Yang, Bingyi Kang, Zilong Huang, Zhen Zhao, Xiaogang Xu, Jiashi Feng, and Hengshuang Zhao.
\newblock Depth anything v2.
\newblock \emph{Advances in Neural Information Processing Systems}, 37:\penalty0 21875--21911, 2024{\natexlab{b}}.

\bibitem[Ye and Bilodeau(2022)]{ye2022vptr}
Xi Ye and Guillaume-Alexandre Bilodeau.
\newblock Vptr: Efficient transformers for video prediction.
\newblock In \emph{2022 26th International Conference on Pattern Recognition (ICPR)}, pages 3492--3499. IEEE, 2022.

\bibitem[Ye and Bilodeau(2023)]{ye2023unified}
Xi Ye and Guillaume-Alexandre Bilodeau.
\newblock A unified model for continuous conditional video prediction.
\newblock In \emph{Proceedings of the IEEE/CVF Conference on Computer Vision and Pattern Recognition (CVPR) Workshops}, pages 3603--3612, 2023.

\bibitem[Ye and Bilodeau(2024)]{STDiff}
Xi Ye and Guillaume-Alexandre Bilodeau.
\newblock Stdiff: Spatio-temporal diffusion for continuous stochastic video prediction.
\newblock In \emph{Proceedings of the AAAI Conference on Artificial Intelligence}, pages 6666--6674, 2024.

\bibitem[Yu et~al.(2023)Yu, Sohn, Kim, and Shin]{PVDM}
Sihyun Yu, Kihyuk Sohn, Subin Kim, and Jinwoo Shin.
\newblock Video probabilistic diffusion models in projected latent space.
\newblock In \emph{Proceedings of the IEEE/CVF Conference on Computer Vision and Pattern Recognition}, 2023.

\bibitem[Zhai et~al.(2024)Zhai, Lin, Li, Lin, Wang, Yang, Doermann, Yuan, Liu, and Wang]{zhai2024idol}
Yuanhao Zhai, Kevin Lin, Linjie Li, Chung-Ching Lin, Jianfeng Wang, Zhengyuan Yang, David Doermann, Junsong Yuan, Zicheng Liu, and Lijuan Wang.
\newblock Idol: Unified dual-modal latent diffusion for human-centric joint video-depth generation.
\newblock In \emph{Proceedings of the European Conference on Computer Vision}, 2024.

\bibitem[Zhang et~al.(2024{\natexlab{a}})Zhang, Wei, Jiang, ZHANG, Zuo, and Tian]{zhang2023controlvideo}
Yabo Zhang, Yuxiang Wei, Dongsheng Jiang, XIAOPENG ZHANG, Wangmeng Zuo, and Qi Tian.
\newblock Controlvideo: Training-free controllable text-to-video generation.
\newblock In \emph{The Twelfth International Conference on Learning Representations}, 2024{\natexlab{a}}.

\bibitem[Zhang et~al.(2024{\natexlab{b}})Zhang, Hu, Cheng, Paudel, and Yang]{ExtDM}
Zhicheng Zhang, Junyao Hu, Wentao Cheng, Danda Paudel, and Jufeng Yang.
\newblock Extdm: Distribution extrapolation diffusion model for video prediction.
\newblock In \emph{Proceedings of the IEEE/CVF International Conference on Computer Vision (CVPR)}, 2024{\natexlab{b}}.

\end{thebibliography}
}

\clearpage
\setcounter{page}{1}
\maketitlesupplementary
\section{Further implementation details}
\subsection{Latent autoencoders}
The autoencoders are trained separately on each dataset following the pipeline defined by the authors of PVDM~\cite{PVDM}. We train them until convergence of FVD, SSIM and PSNR and proceed with a second fine-tuning stage with the adversarial loss for very few iterations.
Given a sequence of frames $\mathbf{x}$ of shape $T\times H\times W\times C$ as input, the encoder produces a latent vector of shape $C'\times L$, where $L$ is computed as follows: 
$\frac{H\cdot W}{P^2}+\frac{T}{P}\cdot (H+W)$.
We use the default configuration of PVDM and keep the patch size $P=4$ for $64\times64$ and $128\times128$ resolution, while we use $P=8$ for higher resolutions.
The number of channels for the latent vector is also set to the default value $C'=4$ for all experiments up to $128\times128$ resolution, otherwise we use $C'=16$.
For example, for Cityscapes ($128\times128$) the latent vector has the shape of 
$4 \times 1536$.
The number of hidden channels for the autoencoder is set to 192, differing from 384 used in the original version for efficiency reasons.

\subsection{BAIR depth ground-truth}\label{bair_depth_compute}
We generate ground-truth depth images for the BAIR~\cite{BAIR} dataset using the off-the-shelf DepthAnything-v2~\cite{depth_anything_v2} model, specifically the `\texttt{vit-b}' version. Depth estimation is performed frame by frame, which can result in flickering in the depth videos due to the lack of temporal coherence.

\subsection{SYNTHIA semantic segmentation maps}
The SYNTHIA dataset~\cite{SYNTHIA} provides semantic segmentation maps for each RGB frame, consisting of 16 distinct classes. These maps are represented as 3-channel images, where the first channel encodes the class IDs, and the remaining two channels assign instance IDs to individual objects. For our work, we focused solely on the class IDs, transforming the first channel into a grayscale image by dividing the original values by 15 and rescaling them to the range $[0, 255]$. To resize the images to $128\times128$ pixels, we employed nearest-neighbor down-sampling in order to preserve the integrity of class labels and avoid blending between classes during resizing.

\subsection{ERA5-Land data processing}
Surface pressure (sp) is expressed in Pascals (Pa), while the two-meter temperature (t2m) is given in Kelvin (K). To adapt these data modalities to our training setup, we normalized them using the minimum and maximum values provided in the dataset's metadata. For evaluation, we rescale the predictions to their original range and compute the $L_1$ error. The data contains NaN values where the measurements are missing (e.g.\ seas and ocean). We set these values to 0 for training and mask them out in the prediction before computing evaluation metrics.

\subsection{OpenDV-YouTube training}
We additionally experimented on higher resolution videos ($256\times256$). Specifically, we re-trained our approach on Cityscapes in a $8\rightarrow8$ setting with $256\times256$ resolution.
We then finetune this model on a small subset of the OpenDV-Youtube~\cite{yang2024genad} dataset. 
The subset essentially includes one video (ID: \texttt{JS0gJxhFFJ8}) of 65 minutes at 30 fps. The initial and last frames containing text overlay are dropped.
The depth maps for these videos are not available, thus, similarly to BAIR~(\cref{bair_depth_compute}), we used DepthAnything-v2~\cite{depth_anything_v2} version `\texttt{vit-l}' on the raw images before center cropping and resizing them to $256\times256$.
We report the evaluation metrics in~\cref{tab:opendv_results} and show additional qualitative results in~\cref{fig:youtube}.
\begin{table}[h!]
\centering
\resizebox{0.8\linewidth}{!}{%
\begin{tabular}{lccc|cc}
\toprule
\multirow{2}{*}{Models} & \multicolumn{3}{c}{RGB} & \multicolumn{2}{|c}{Depth} \\
& FVD$\downarrow$ & SSIM$\uparrow$ & LPIPS$\downarrow$ & SSIM$\uparrow$ & $L_{2}\downarrow$\\
\midrule
SyncVP & 247.14 & 0.611 & 224.17 & 0.972 & 1.1703\\
\bottomrule
\end{tabular}
}

\caption{Results on OpenDV-Youtube ($256\times256$, $8\rightarrow24$).}
\label{tab:opendv_results}
\end{table}
\section{Training ablation}
We provide additional ablation results~(\cref{tab:two-stage-ablation}) to show the benefits of using a two-stage training pipeline. In the first stage we learn $p(\mathbf{r}_x \mid \mathbf{r}_c)$ and $p(\mathbf{d}_x \mid \mathbf{d}_c)$, while we exploit in the second stage these pre-trained weights to learn the joint conditional distribution $p(\mathbf{r}_x,\mathbf{d}_x\mid \mathbf{r}_c,\mathbf{d}_c)$. To evaluate this, we compare the results of our SyncVP with a version of the model trained from scratch directly on multi-modal data.

\begin{table}
\centering
\resizebox{1.0\linewidth}{!}{%
\begin{tabular}{c|ccc|cc}
\toprule
\multirow{2}{*}{\shortstack{Two-stage\\training}} & \multicolumn{3}{c}{RGB} & \multicolumn{2}{|c}{Depth} \\
& FVD$\downarrow$ & SSIM$\uparrow$ & LPIPS$\downarrow$ & SSIM$\uparrow$ & $L_{2}\downarrow$\\
\midrule
\ding{55} & 158.53 & \textbf{0.674} & 176.45 & 0.827 & 8.044\\
\checkmark & \textbf{84} & 0.649 & \textbf{159.73} & \textbf{0.830} & \textbf{7.329}\\
\bottomrule
\end{tabular}

}
\caption{Ablation on Cityscapes about the impact of the proposed two-stage training pipeline.}
\label{tab:two-stage-ablation}
\end{table}

\begin{figure}[h!]
  \centering \includegraphics[width=0.8\linewidth]{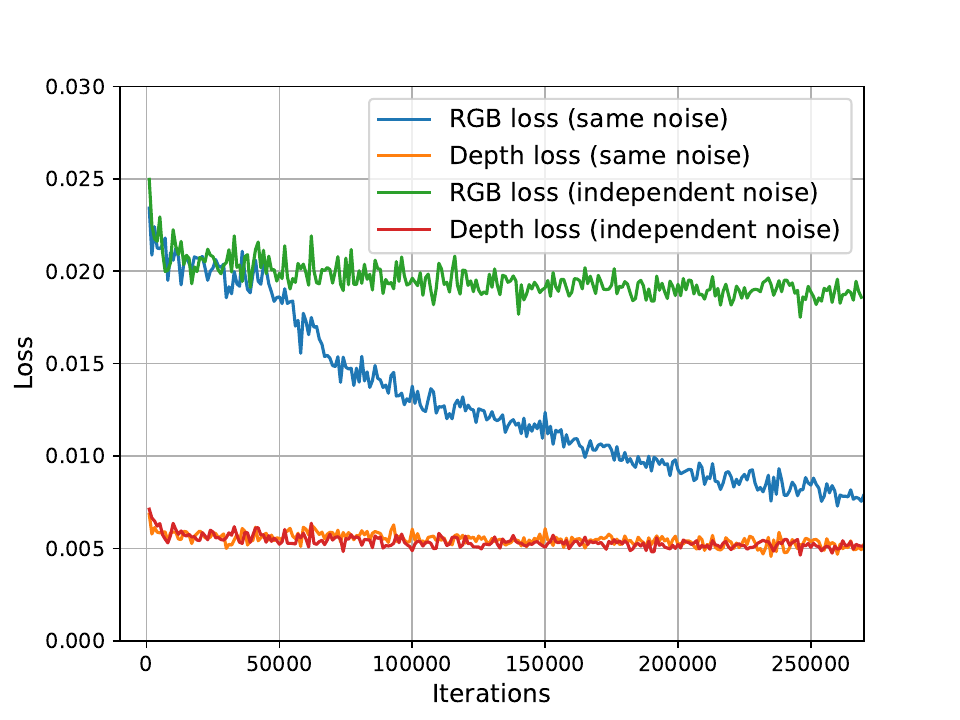}
   \caption{Training loss comparison between using the same noise for both modalities or independent noise
for each modality.}
   \label{fig:same_vs_ind_losses}
\end{figure}

In~\cref{fig:same_vs_ind_losses}, we show the loss plot to further validate the effectiveness of our shared noise strategy during training.

\section{Additional qualitative results}
\begin{figure*}[t!]
  \begin{subfigure}[t]{\linewidth}
  \centering
    \includegraphics[width=\linewidth]{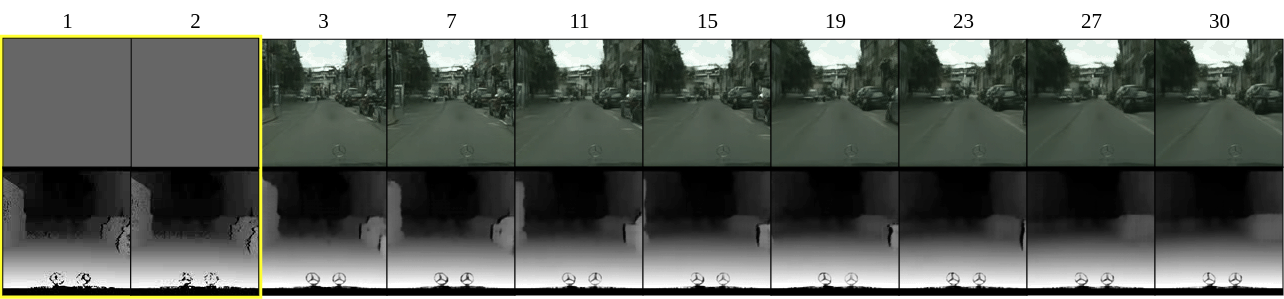}
    \caption{Prediction on Cityscapes using only depth conditioning.}
    \label{fig:city_pred_without_rgb}
  \end{subfigure}
  \begin{subfigure}[t]{\linewidth}
  \centering
    \includegraphics[width=\linewidth]{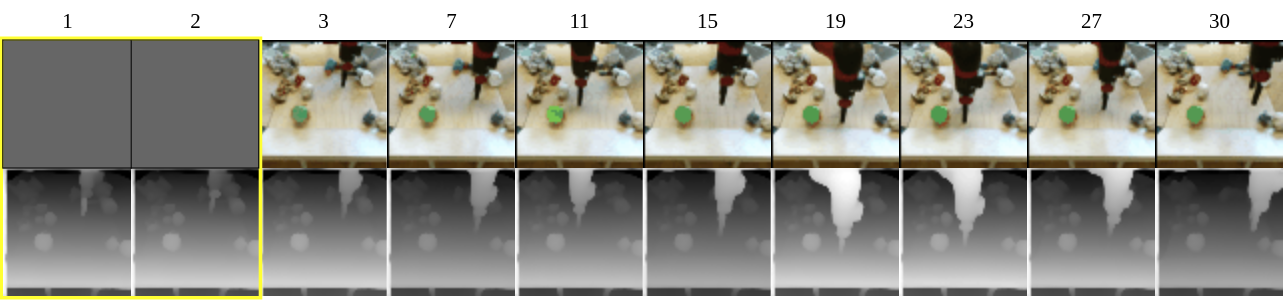}
    \caption{Prediction on BAIR using only depth conditioning.}
    \label{fig:bair_pred_without_rgb}
  \end{subfigure}
  \begin{subfigure}[t]{\linewidth}
  \centering
    \includegraphics[width=\linewidth]{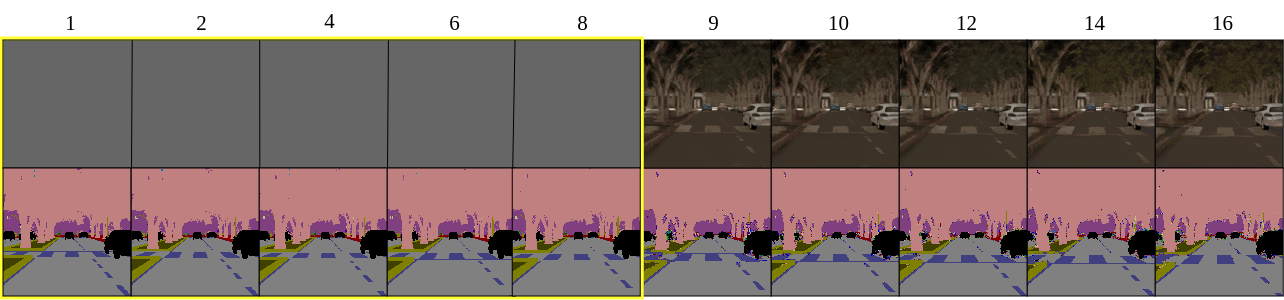}
    \caption{Prediction on SYNTHIA using only semantic segmentation maps conditioning.}
    \label{fig:syn_pred_without_rgb}
  \end{subfigure}
  \caption{SyncVP video prediction without conditioning on past RGB frames .}
  \label{fig:suppl_no_rgb_cond}
\end{figure*}

\begin{figure*}[t!]
  \begin{subfigure}[t]{\linewidth}
  \centering
    \includegraphics[width=\linewidth]{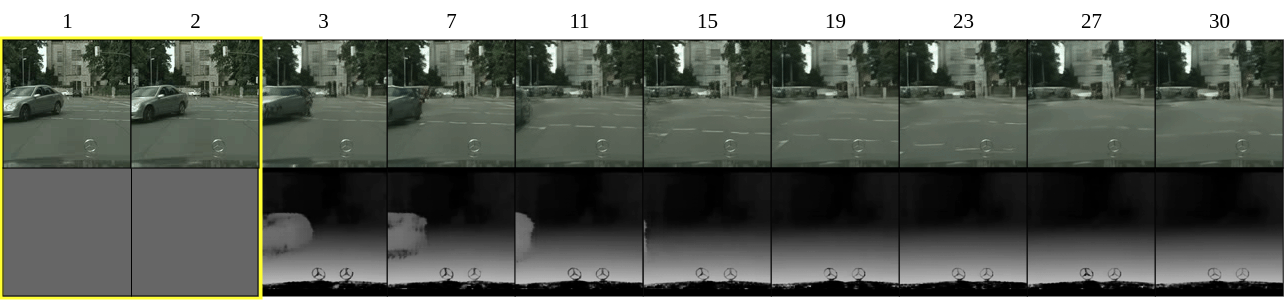}
    \caption{Prediction on Cityscapes.}
    \label{fig:city_pred_without_depth}
  \end{subfigure}
  \begin{subfigure}[t]{\linewidth}
  \centering
    \includegraphics[width=\linewidth]{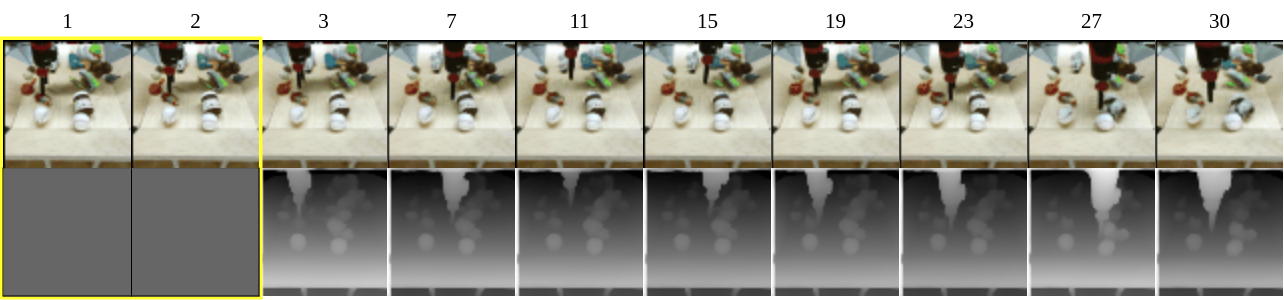}
    \caption{Prediction on BAIR.}
    \label{fig:bair_pred_without_depth}
  \end{subfigure}
  \begin{subfigure}[t]{\linewidth}
  \centering
    \includegraphics[width=\linewidth]{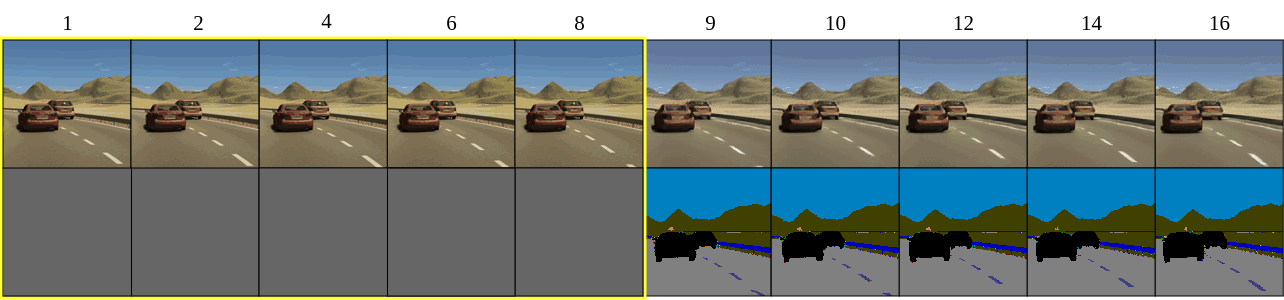}
    \caption{Prediction on SYNTHIA.}
    \label{fig:syn_pred_without_sem}
  \end{subfigure}
  \caption{SyncVP video prediction with conditioning only on RGB frames.}
  \label{fig:suppl_no_depth_cond}
\end{figure*}

We provide some more qualitative results, particularly for cases where one modality is missing. Specifically, \cref{fig:suppl_no_rgb_cond} demonstrates how our model is able to predict future frames using only the non-RGB modality as observation. Such task is way more complex than standard joint conditional generation or the case in which the low detail modality (depth or semantic) is missing.
Nevertheless, the results still exhibit strong cross-modal alignment between the predicted frames.
\cref{fig:suppl_no_depth_cond} shows further examples where the model is conditioned solely on past RGB frames. In these cases, our approach is still able to predict aligned depth or semantic segmentation images.

\subsection{Agriculture data}
We additionally apply our model on agricultural data collected from a sweet pepper greenhouse. The dataset contains images of 415 plants captured on four different dates, which can be considered as 415 short video sequences. Given the dataset’s limited size and the challenge of forecasting the plant growth stages, we adopt a leave-one-out strategy for training and testing. 
All images are resized to a resolution of $512\times288$, and depth is estimated using DepthAnything-v2~\cite{depth_anything_v2}. An example of a prediction is shown in~\cref{fig:agriculture_data}.

\section{Robustness to noise}
\begin{table}[h!]
\centering
\resizebox{.6\linewidth}{!}{%
\begin{tabular}{lccc}
\toprule
{noise} & \multicolumn{3}{c}{RGB} \\
$\sigma$ & FVD$\downarrow$ & SSIM$\uparrow$ & LPIPS$\downarrow$\\
\midrule
5 & 87.78 & 0.641 & 161.61 \\
2.5 & 85.98 & 0.644 & 160.75\\
0 & \textbf{84} & \textbf{0.649} & \textbf{159.73}\\
\bottomrule
\end{tabular}
}
\caption{Impact of Gaussian noise in observed disparity (Cityscapes).}
\label{tab:noise_levels_input}
\end{table}
Beside showing the ability of SyncVP to deal with missing modality input, we further test its resilience to noisy input. 
Namely, we inject random Gaussian noise with increasing $\sigma$ on the depth input.
As shown in \cref{tab:noise_levels_input}, the noisy input does not affect much the predictions.

\begin{figure*}[t!]
  \centering
    \includegraphics[width=\linewidth]{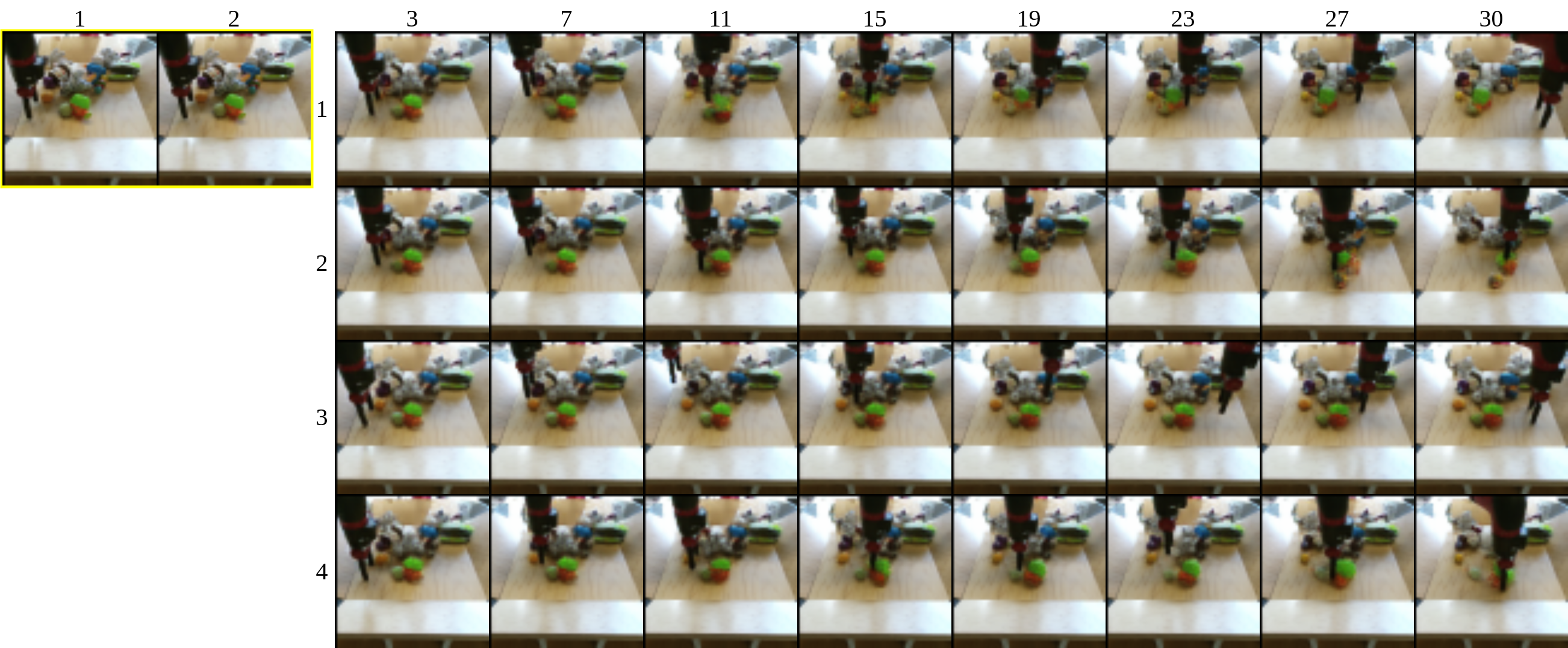}
    \caption{Multiple predicted trajectories for the same observation on BAIR.}
    \label{fig:bair_multi_traj}
\end{figure*}
\begin{figure*}[t!]
  \centering
    \includegraphics[width=\linewidth]{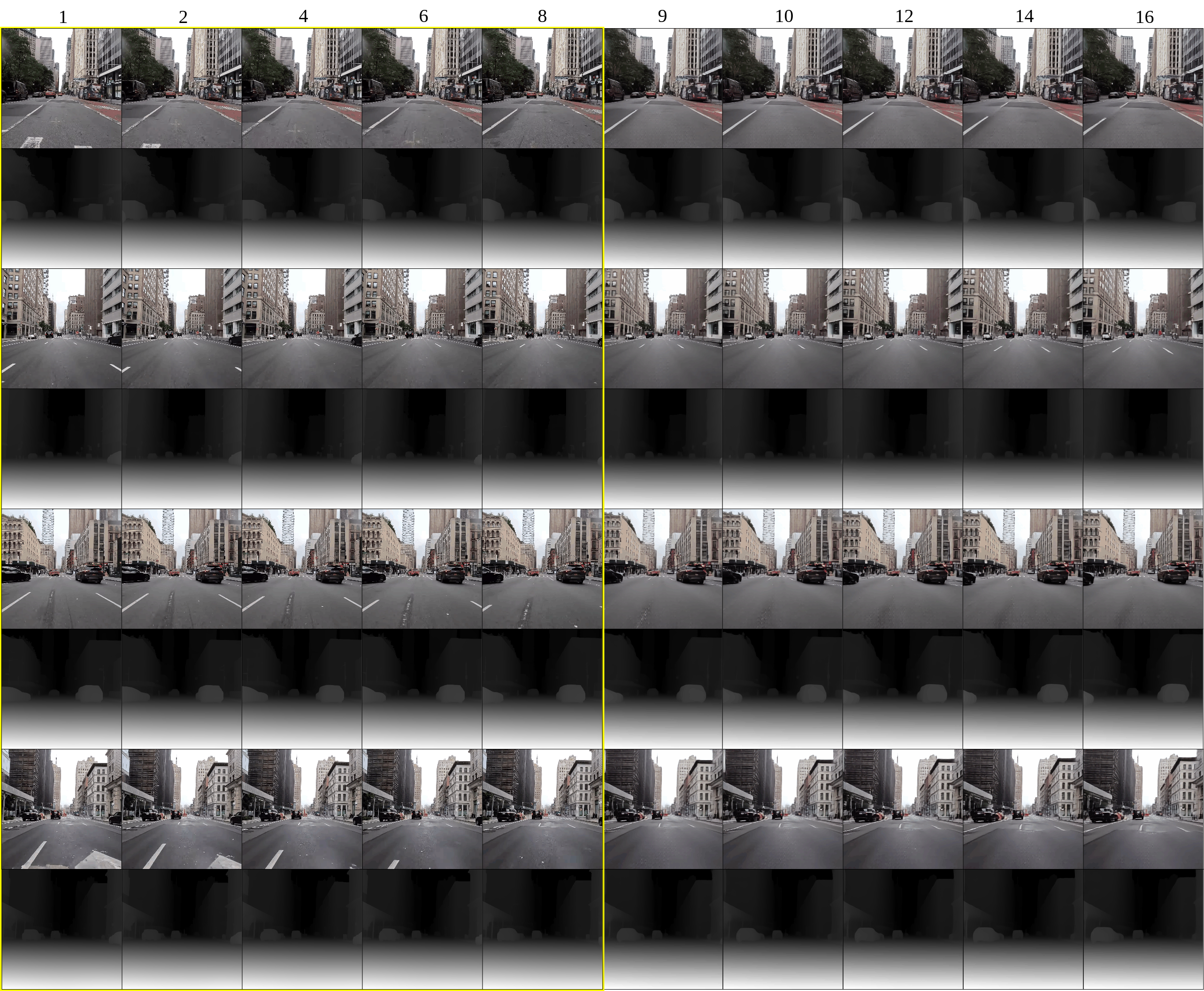}
    \caption{
    SyncVP predictions on OpenDV-Youtube~\cite{yang2024genad} ($256\times256$).
    }
    \label{fig:youtube}
\end{figure*}

\begin{figure*}[t!]
  \centering
    \includegraphics[width=\linewidth]{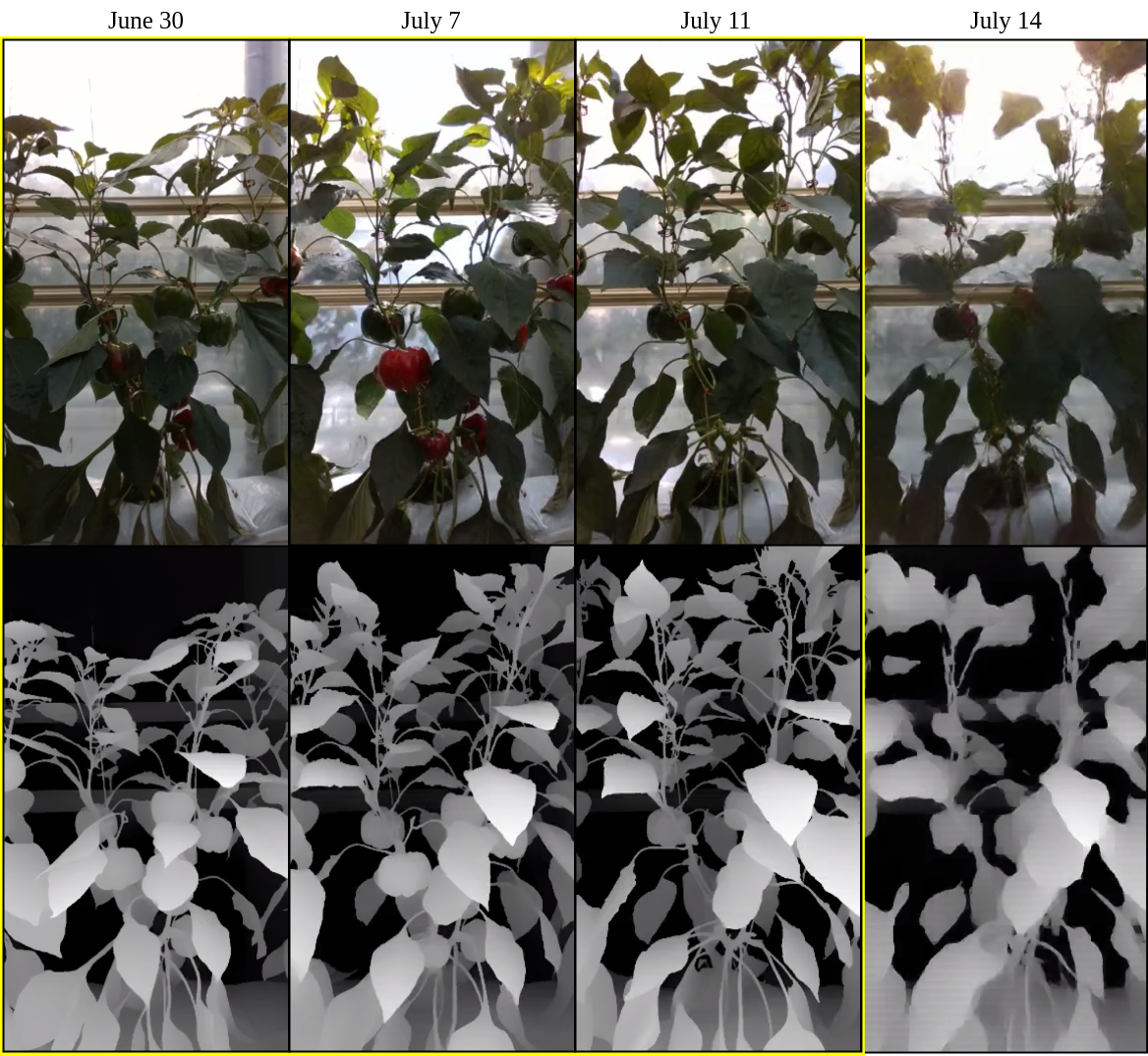}
    \caption{SyncVP prediction ($3\rightarrow1$) on agriculture timeseries data.}
    \label{fig:agriculture_data}
\end{figure*}

\end{document}